\documentclass{article}

\PassOptionsToPackage{numbers, compress}{natbib}

 \usepackage[preprint]{neurips_2026}

\usepackage[utf8]{inputenc} % allow utf-8 input
\usepackage[T1]{fontenc}    % use 8-bit T1 fonts
\usepackage{hyperref}       % hyperlinks
\usepackage{url}            % simple URL typesetting
\usepackage{booktabs}       % professional-quality tables
\usepackage{amsfonts}       % blackboard math symbols
\usepackage{nicefrac}       % compact symbols for 1/2, etc.
\usepackage{microtype}      % microtypography
\usepackage{xcolor}         % colors

\usepackage{amsmath,amsfonts,bm}

\def\eqref#1{equation~\ref{#1}}
\def\1{\bm{1}}

\DeclareMathAlphabet{\mathsfit}{\encodingdefault}{\sfdefault}{m}{sl}
\SetMathAlphabet{\mathsfit}{bold}{\encodingdefault}{\sfdefault}{bx}{n}

\usepackage{graphicx}
\usepackage{subcaption} 
\usepackage{epstopdf}
\usepackage{algorithm}
\usepackage{algorithmic}
\usepackage{wrapfig}

\usepackage{amsmath}
\usepackage[normalem]{ulem}
\usepackage{pgfplots}

\usepackage{tikz}
\usepackage{tikz-3dplot}

\usepackage[inline]{asymptote}
\usepackage[export]{adjustbox}
\usepackage{lmodern}
\usepackage{xspace}
\usepackage{pdfpages}
\usepackage{listings}
\usepackage{import}

\usepackage{microtype}

\usepackage{amsmath}
\usepackage{amssymb}
\usepackage{mathtools}
\usepackage{amsthm}

\usepackage{dblfloatfix}
\usepackage{placeins}  

\theoremstyle{plain}

\theoremstyle{definition}

\theoremstyle{remark}

\usepackage[capitalize,noabbrev]{cleveref}
\crefname{fig.}{Fig.}{Figs.}
\Crefname{figure}{Figure}{Figures}
\crefname{tab.}{Tab.}{Tabs.}
\Crefname{table}{Table}{Tables}
\crefname{eq.}{Eq.}{Eqs.}
\Crefname{equation}{Equation}{Equations}
\crefname{sec.}{Sec.}{Secs.}
\Crefname{section}{Section}{Sections}
\crefname{appendix}{Appendix}{Appendices}
\Crefname{appendix}{Appendix}{Appendices}
\newcommand{\ie}{i.e.\xspace}
\newcommand{\eg}{e.g.\xspace}

\newcommand{\clipscore}{CLIPScore\xspace}
\newcommand{\gpt}{GPT-4V\xspace}

\usetikzlibrary{patterns,shapes.geometric,calc}
\usepgfplotslibrary{groupplots}
\usetikzlibrary{pgfplots.groupplots}
\usetikzlibrary{calc,decorations.pathmorphing}
\usetikzlibrary{backgrounds}
\pgfplotsset{compat=1.18}
\usepgfplotslibrary{polar}
\usepgfplotslibrary{fillbetween}

\definecolor{pastel_green}{RGB}{218, 247, 166}
\definecolor{pastel_yellow}{RGB}{255, 195, 0}
\definecolor{pastel_red}{RGB}{217, 122, 136}
\definecolor{pastelorange}{RGB}{255, 218, 180}
\definecolor{pastelpink}{RGB}{255, 203, 225}
\definecolor{pastelblue}{RGB}{174,198,207}
\definecolor{teagreen}{RGB}{214, 229, 189}
\definecolor{pastelpurple}{RGB}{202, 191, 255}

\definecolor{pastel1}{RGB}{141, 211, 199}
\definecolor{pastel2}{RGB}{249, 225, 168}
\definecolor{pastel3}{RGB}{220, 204, 236}
\definecolor{pastel4}{RGB}{251, 128, 114}
\definecolor{pastel5}{RGB}{128, 177, 211}
\definecolor{pastel6}{RGB}{255, 245, 157}

\useunder{\uline}{\ul}{}

\definecolor{codegreen}{rgb}{0,0.6,0}
\definecolor{codegray}{rgb}{0.5,0.5,0.5}
\definecolor{codepurple}{rgb}{0.58,0,0.82}
\definecolor{backcolour}{rgb}{0.95,0.95,0.92}

\lstdefinestyle{mystyle}{
    backgroundcolor=\color{backcolour},   
    commentstyle=\color{codegreen},
    keywordstyle=\color{magenta},
    numberstyle=\tiny\color{codegray},
    stringstyle=\color{codepurple},
    basicstyle=\ttfamily\footnotesize,
    breakatwhitespace=false,         
    breaklines=true,                 
    captionpos=b,                    
    keepspaces=true,                 
    numbers=left,                    
    numbersep=5pt,                  
    showspaces=false,                
    showstringspaces=false,
    showtabs=false,                  
    tabsize=2
}

\title{LoRAtorio: An intrinsic approach to LoRA  
  \\ Skill Composition}

\author{%
  Niki Foteinopoulou$^{1}$\thanks{Corresponding authors: \texttt{niki.foteinopoulou@toshiba.eu}, \texttt{stephan.liwicki@toshiba.eu}}
  \quad Ignas Budvytis$^{2}$
  \quad Stephan Liwicki$^{1}$\footnotemark[1] \\
  $^{1}$Cambridge Research Laboratory, Toshiba Europe \\ \quad
  $^{2}$Independent Researcher \\
}

\begin{document}

\maketitle

\begin{abstract}
Low-Rank Adaptation (LoRA) has become a widely adopted technique in text-to-image diffusion models, enabling the personalisation of visual concepts such as characters, styles, and objects. However, existing approaches struggle to effectively compose multiple LoRA adapters, particularly in open-ended settings where the number and nature of required skills are not known in advance. In this work, we present LoRAtorio, a novel train-free framework for multi-LoRA composition that gates each adapter using only signals already inside the denoiser, specifically, patch-level cosine (dis)similarity between each LoRA's predicted noise and the base model's. We refer to this as ``intrinsic'' guidance because no external parameters are introduced. Our method is motivated by two key observations: (1) LoRA adapters trained on narrow domains produce unconditioned denoised outputs that diverge from the base model, and (2) when conditioned out of distribution, LoRA outputs show behaviour closer to the base model than when conditioned in distribution.
These patch-level similarities are used to construct a spatially-aware weight matrix, which guides a weighted aggregation of LoRA outputs. To address domain drift, we further propose a modification to classifier-free guidance that incorporates the base model's unconditional score into the composition. We extend this formulation to a dynamic module selection setting, enabling inference-time selection of relevant LoRA adapters from a large pool. On the \textit{ComposLoRA} benchmark, LoRAtorio achieves state-of-the-art performance, with \clipscore that does not deteriorate as the number of composed adapters grows, surpassing the strongest prior method by up to 1.3\% in \clipscore, and reaching a 76.92\% \gpt win rate against the closest baseline. We further show that the framework is backbone-agnostic: the intrinsic-similarity principle transfers to rectified-flow architectures (FLUX.1-dev) without retraining. Project page: \url{https://nickyfot.github.io/loratorio-skill-composer/}.
\end{abstract}

\section{Introduction}
Diffusion models operate by gradually learning to reverse a noise process, effectively capturing the underlying data distribution through iterative denoising~\citep{ho2020denoising, nichol2021improved, song2021scorebased}. In practice, this enables them to approximate the complex structure of their training data and generate new, previously unseen samples that remain faithful to the original data's domain. Beyond base text-to-image generation capabilities, works such as Dreambooth~\citep{ruiz2023dreambooth} and StyleDrop~\citep{sohn2023styledrop} have enabled personalisation and fine-grained customisation. These approaches often rely on LoRA adapters~\citep{hu2022lora}, which specialise a base model to preserve the identity of specific concepts or objects, supporting applications like virtual try-on~\citep{lobba2025inverse} and avatar generation~\citep{Huang_2024_CVPR}. Each LoRA adapter effectively encodes a ``skill'' or concept, and generation with a single adapter yields precise, high-quality outputs. However, when multiple skills are loaded simultaneously into a single model instance, we observe a rapid deterioration in performance~\citep{zhongmulti2024, prabhakar2025lora}. Understanding the source of this degradation is key to enabling reliable multi-concept generation.

\begin{figure}[tbp]
    \centering
    % First subfigure
    \begin{subfigure}[b]{0.46\textwidth}
        \centering 
        \includegraphics[width=\textwidth]{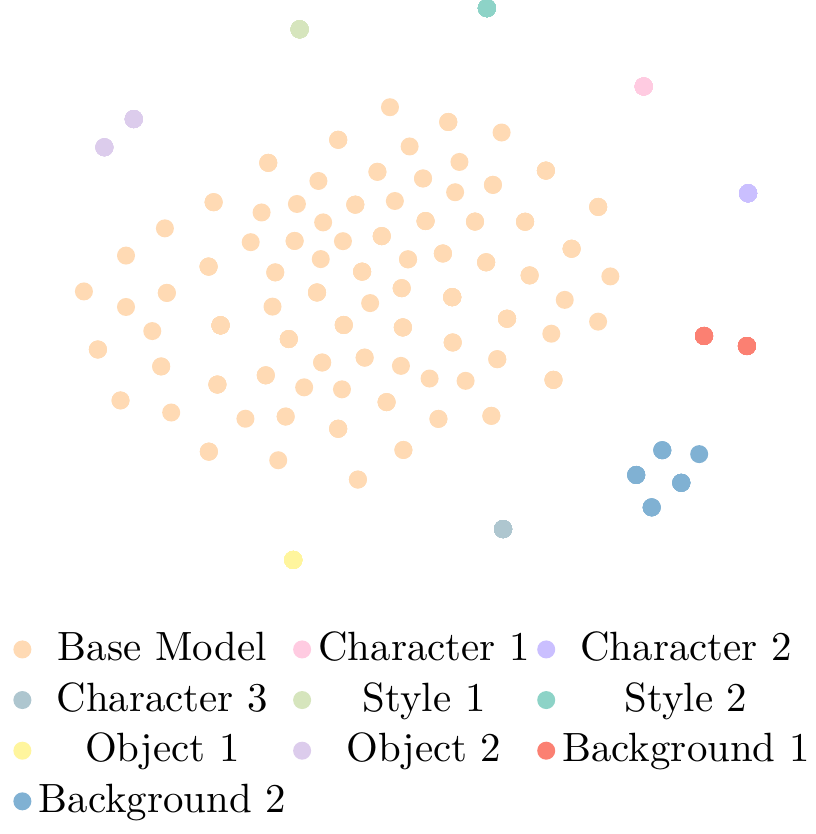}
        \caption{}
        \label{fig:basemodelnoise}
    \end{subfigure}\hfill
    % Second subfigure
    \begin{subfigure}[b]{0.52\textwidth}
        \centering
        \includegraphics[width=\textwidth]{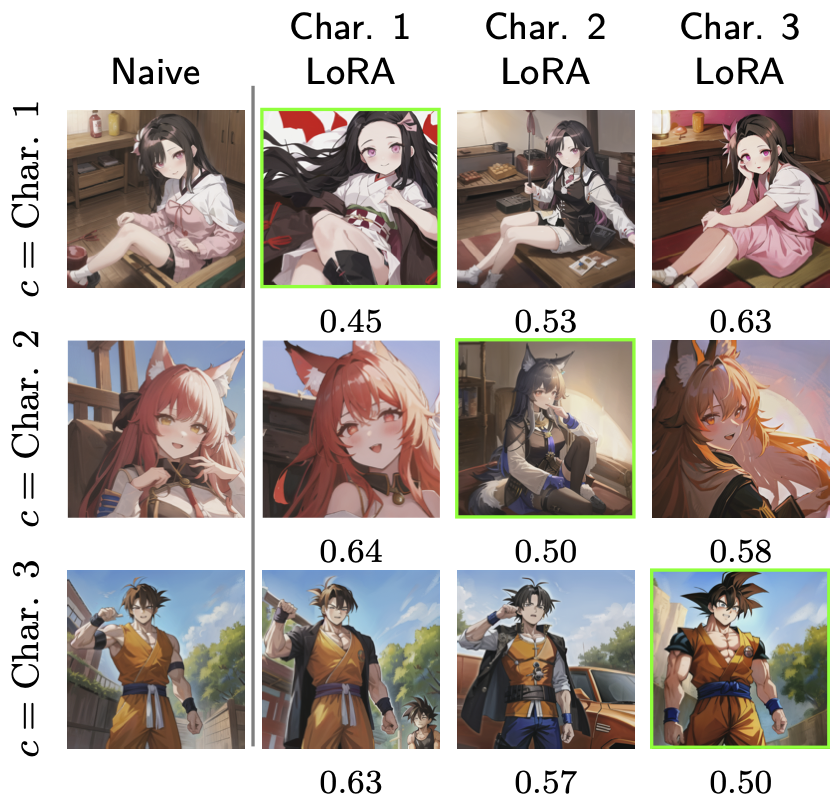}
        \caption{}
        \label{fig:wrongpromptlora}
    \end{subfigure}
    
    \caption{Observations of noise prediction from the \textit{ComposLoRA} testbed. (a) t-SNE visualisation of the unconditioned latent space, for the Base Model and LoRA-adapted models. (b) Generated images from each Character LoRA, conditioned with text prompts originally associated with other LoRAs. When a prompt falls outside a LoRA’s training distribution, the predicted latent $e_{\theta_{i}}(z_0, 0, c)$ of the $i^{th}$ LoRA tends to align closely with that of the base (Na\"{\i}ve) model, showing minimal deviation due to changes in $p_{\theta_i}(x)$, also shown by the cosine similarity of the conditioned noise prediction of the Na\"{\i}ve model with that of the $i^{th}$ LoRA.}

    \label{fig:combined_lora_results}
\end{figure}

To better understand the challenges of composing multiple LoRA adapters, we begin with a preliminary analysis of their behaviour. Specifically, we examine the unconditional noise representations produced by the base model and various LoRA-augmented models. We observe that the distribution of the LoRA diverges from that of the base model (\cref{fig:basemodelnoise}), particularly when LoRAs are trained on narrow or highly specialised datasets, conditions common in personalisation settings~\citep{li2024lorasc, ruiz2023dreambooth}. This domain shift also manifests in conditioned outputs, as evident through visual inspection in~\cref{fig:wrongpromptlora} (details in~\cref{sec:kl_divergence}).

\textbf{Observation 1:} 
The unconditioned noise estimate $e_i(z, t)$ produced by the $i^\text{th}$ LoRA differs from that of the
base model $e(z, t)$.

While this divergence is notable, especially under unconditional or in-domain conditions, LoRA adapters are also known to mitigate catastrophic forgetting, particularly compared to fully fine-tuned models~\citep{biderman2024lora}. That is, LoRAs tend to preserve the base model’s generalisation capabilities. Indeed, we observe that even though there are stylistic changes in the generated image as a result of the loaded LoRAs, the composition and theme of the generated output more closely resemble the base model when a text condition outside the LoRA distribution is given. This is attributed to the sparse and low-norm nature of LoRA weights~\citep{fu_effectiveness_2023, shah2024ziplora}. To quantify this effect, we measure cosine similarities between the noise scores of LoRA-augmented models and the base model, both within and outside LoRA’s training distribution. These measurements show high similarity to the base model out-of-distribution and lower similarity in-distribution, which, together with the visual inspection in \cref{fig:wrongpromptlora}, support the following:

\textbf{Observation 2:} When the input condition lies outside the LoRA’s target domain, the output of the augmented model more closely resembles that of the base model.

Previous skill composition research utilises \textbf{trainable} methods, such as mixture-of-experts or hyper-networks~\citep{charakorn_textlora_2025, shenaj_lorarar_2024, zhu_mole_2024}, and \textbf{train-free} approaches involving scheduling, latent merging, or weight arithmetic~\citep{zhongmulti2024, zou_cached_2025, li2024selma, yang2024lora, huggingface2024merge, huang2023lorahub, shah2024ziplora}. Trainable methods often prove impractical due to the high cost of retraining for new concepts and data confidentiality constraints. Conversely, existing train-free techniques, relying on manual schedules or unweighted score aggregation, suffer from performance degradation as the number of skills increases~\citep{prabhakar2025lora, zhongmulti2024}. Critically, all prior works assume the required LoRAs are pre-specified. This is unfeasible in real-world settings like personalised advertising, where user intent is only known at inference time, and the combinatorial space of skill mixtures is vast.

Motivated by Observations~1 and~2, we propose LoRAtorio, a train-free method that gates LoRA contributions using their patch-level divergence from the base model in latent space. Per-patch divergence is converted via a SoftMin into spatially-aware gating weights, so each LoRA contributes most where it differs most from the base; a re-centered classifier-free guidance further corrects for unconditional drift by mixing in the base model's unconditional score. Unlike prior approaches that rely on extrinsic signals such as RGB frequency~\citep{zou_cached_2025} or hand-crafted scheduling~\citep{zhongmulti2024}, LoRAtorio is entirely intrinsic; the gating signal is the consistency between LoRA and base representations themselves. We further extend the task to a \emph{dynamic} module-selection setting, where all available adapters are loaded, and the relevant subset is chosen at inference time, reflecting real-world scenarios in which the required skills are not known a priori.

Our main contributions can be summarised as follows:
\begin{itemize}
    \item We introduce \textbf{LoRAtorio}, a train-free method that aggregates multiple LoRA adapters using a patch-level dissimilarity score between each LoRA's predicted noise and the base model's, the first composition mechanism gated by an intrinsic, training-free metric.
    \item Furthermore, we propose a modification to classifier-free guidance that re-centers the unconditional score on the base model (\cref{sec:recenter},~\cref{fig:method_recenter}), correcting the domain drift implied by Observation~1.
    \item We further introduce a dynamic module-selection regime in which a candidate pool of LoRAs is loaded simultaneously and the relevant subset is chosen at each denoising step from the same intrinsic similarity signal, removing the need for prompt-time scheduling.
\end{itemize}
We demonstrate that LoRAtorio achieves state-of-the-art (SoTA) performance on the \textit{ComposLoRA} benchmark both in terms of automated metrics and human preference. This is consistent for both static and dynamic module settings. As a further extension demonstrating the generality of the intrinsic-similarity principle, we adapt the framework to a rectified-flow~\citep{esser2024scaling} backbone (FLUX.1-dev), the first spatially-aware multi-LoRA composition method on rectified-flow backbones.

\section{LoRAtorio}

\begin{figure*}[t]
    \centering
    \includegraphics[width=\textwidth]{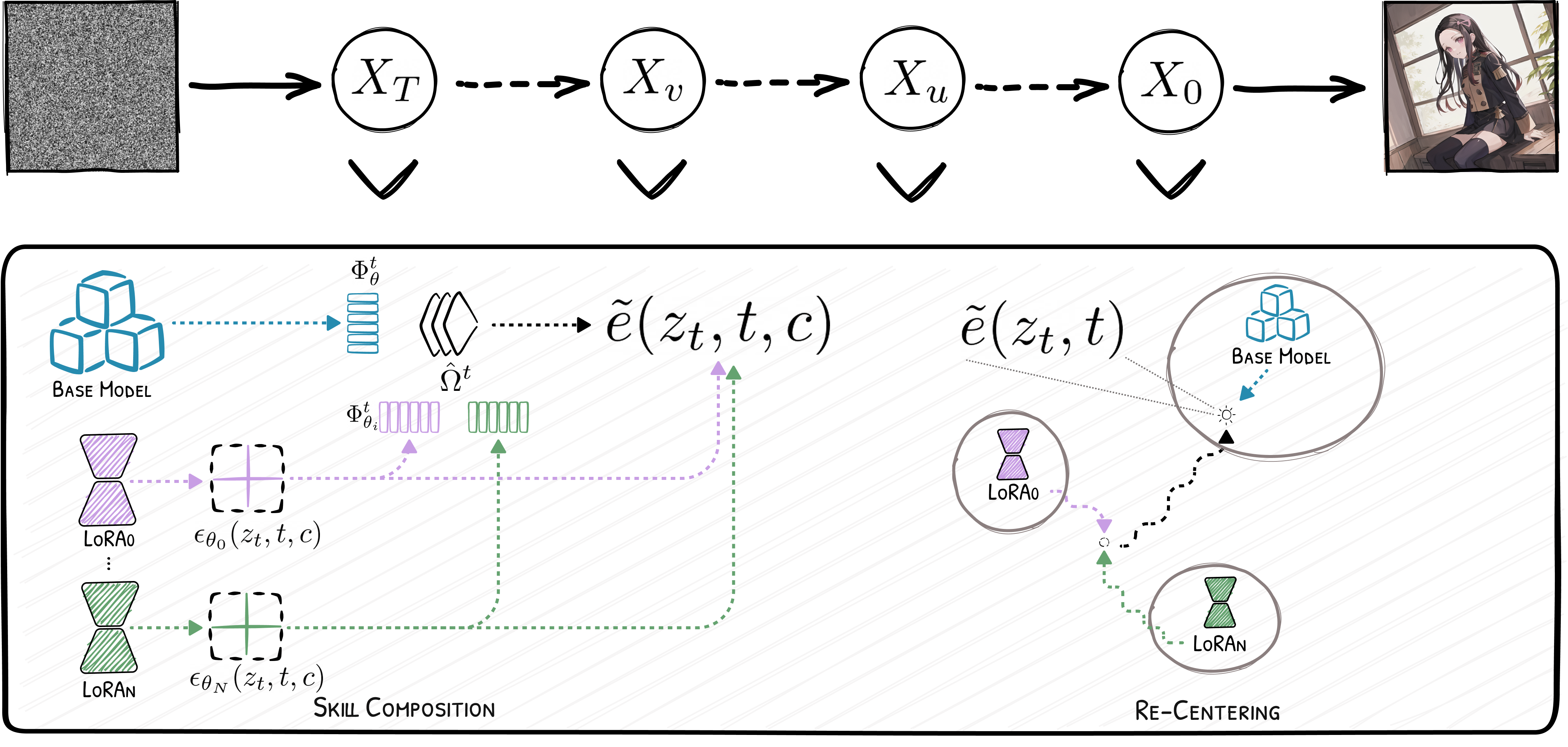}
    \caption{Overview of \textsc{LoRAtorio}. \textbf{Skill Composition:} At each denoising timestep $t$, the conditional score from the $i^\text{th}$ LoRA, $e_{\theta_i}(z_t, t, c)$, is partitioned into $P$ spatial patches (\eqref{eq:tokenise}). Each patch is flattened and compared to its corresponding patch in the base model’s predicted noise using cosine similarity. The resulting similarity matrix is passed through a SoftMin function (\eqref{eq:softmin}) to produce a weight matrix $\Omega$, assigning higher weights to patches that diverge more from the base model. These weights are used to compute a spatially-aware weighted average across LoRA outputs (\eqref{eq:upscale} and~\eqref{eq:weightedavg}). \textbf{Re-Centering:} To mitigate domain drift of the unconditional score by the multiple LoRAs, we alter classifier-free guidance by incorporating a weighted combination of the base model’s unconditional score and the aggregated LoRA score (\eqref{eq:recenter}). }
    \label{fig:method}
\end{figure*}

Our skill composition (\cref{sec:loratorio}) is motivated by Observation~2 (\cref{fig:wrongpromptlora}), while the re-centering guidance (\cref{sec:recenter}) addresses Observation~1 (\cref{fig:basemodelnoise}).

\textbf{Preliminaries:}
Latent Diffusion Models \citep{rombach_high-resolution_2022} operate by performing the denoising diffusion process in a learned latent space. Given an input \( x_0 \), an encoder \( \mathcal{E} \) maps it to a latent representation \( z_0 = \mathcal{E}(x_0) \); during the diffusion process, Gaussian noise is progressively added to \( z_0 \), thus producing a noisy sequence \( \{z_t\}_{t=1}^T \). The diffusion model learns to approximate the reverse process via a denoising network \( e_\theta(z_t, t, c) \), conditioned on context \( c \). Classifier-free guidance (CFG)~\citep{ho2022classifier} is incorporated by training the model with both conditional and unconditional objectives; guidance is applied by modifying the predicted noise as follows:
\begin{equation}\label{eq:cfg}
    \hat{e}_\theta(z_t, t, c) = e_\theta(z_t, t) + s \cdot \left( e_\theta(z_t, t, c) - e_\theta(z_t, t) \right),
\end{equation}

where \( s \geq 1 \) is the guidance scale and \( e_\theta(z_t, t) \) denotes the unconditional prediction. This approach allows the model to maintain sample diversity while enhancing conditional fidelity without relying on external classifiers.

Low-Rank Adaptation (LoRA) \citep{hu2022lora} is a parameter-efficient fine-tuning method that enables adaptation of large models by injecting trainable low-rank matrices into existing weight layers. In the context of diffusion models, LoRA allows modification of the model's behaviour (\eg, emphasising identity features) without altering the full parameter set, thereby reducing the risk of catastrophic forgetting. Specifically, for a weight matrix \( W \in \mathbb{R}^{d \times k} \), LoRA introduces a trainable update \( \Delta W = A B \), where \( A \in \mathbb{R}^{d \times r} \) and \( B \in \mathbb{R}^{r \times k} \), with \( r \ll \min(d, k) \). This decomposition allows the model to adapt key features—such as identity attributes—by updating only a small number of parameters; however, when multiple LoRA adapters are present, a linear combination of the weights may lead to semantic conflicts and reduced image quality~\citep{huang2023lorahub, zhongmulti2024, zou_cached_2025}. To address issues related to weight manipulation techniques,~\cite {zhongmulti2024} proposes aggregating conditional and unconditional scores using a weighted average,  so that for $N$ LoRAs:
\begin{equation}
% \begin{split}
    \Tilde{e}( z_t, t, c) = \frac{1}{N} \sum_{i=0}^N w_i \cdot [e_{\theta_i}( z_t, t) + s \cdot (e_{\theta_i}( z_t, t, c) - e_{\theta_i}( z_t, t))]
% \end{split}
\end{equation}
\noindent where the weights $w$ are a scalar hyperparameter (set to 1).

\subsection{Skill Composition using intrinsic knowledge}\label{sec:loratorio}

We propose \textbf{LoRAtorio}, a method that activates all LoRAs at each timestep by leveraging the similarity between their noise latent representations and that of the base model. Motivated by Observation 2, we compute the cosine similarity between the output of the model after incorporating the $i^{th}$ LoRA, denoted by $e_{\theta_i}(z_t, t, c)$, and the base model’s output $e_{\theta}(z_t, t, c)$.

Since the conditioned predicted noise $e(z_t, t, c)$ retains spatial structure, we first perform channel-wise averaging to reduce the dimensionality from $H \times W \times C$ to $H \times W$.  We then partition each of these 2D maps into $P = (H \cdot W) / d^2$ non-overlapping patches of size $d \times d$ and flatten each patch into a vector of length $d^2$. Let $\phi(\cdot): \mathbb{R}^{H \times W} \rightarrow \mathbb{R}^{P \times d^2}$ denote this process. We denote the resulting tokenised outputs as:
\begin{equation}
    \Phi_\theta^t = \phi(e_\theta(z_t, t, c)), \quad \Phi_{\theta_i}^t = \phi(e_{\theta_i}(z_t, t, c))    \label{eq:tokenise}
\end{equation}
with $\Phi_\theta^t, \Phi_{\theta_i}^t \in \mathbb{R}^{P \times d^2}$.
For each LoRA $i$, we compute the cosine similarity between the corresponding patch vectors of $\Phi_\theta^t$ and $\Phi_{\theta_i}^t$, resulting in a weight matrix $\Omega^t = \left\langle \Phi_\theta^t, \Phi_{\theta_i}^t \right\rangle_{\mathrm{cos}} \in \mathbb{R}^{N \times P}$ where $N$ is the number of LoRAs. We then apply a SoftMin operation along the $N$ dimension:
\begin{equation}\label{eq:softmin}
% \begin{split}
    \hat{\Omega}^t = \mathrm{softmin}_\tau(\Omega^t), \quad \text{where} \quad \mathrm{softmin}_\tau(x) = \frac{\exp\left(-x_i / \tau\right)}{\sum_{j=1}^N \exp\left(-x_j / \tau\right)}
% \end{split}
\end{equation}

and $\tau > 0$ is the temperature parameter controlling the softness of the SoftMin. This makes the weighting interpretable as a soft attention mechanism, where LoRAs that diverge more from the base model are given stronger influence in regions where they are more confident. Note the use of SoftMin: low cosine similarity, \ie greater divergence from the base model, yields a small input value and therefore a high gating weight, in line with our interpretation of divergence as evidence of in-distribution adapter activity.
We upscale $\hat{\Omega}^t \in \mathbb{R}^{N \times (H/d \cdot W/d)}$ to match the spatial resolution of the original feature map using a Kronecker product:
\begin{equation}
    \hat{\Omega}^{t, \text{up}} = \hat{\Omega}^t \otimes \mathbf{1}_{d \times d} \label{eq:upscale}
\end{equation}
where $\mathbf{1}_{d \times d}$ is a matrix of ones. This operation effectively repeats each similarity value over a $d \times d$ block. The upscaled similarity maps are then used to modulate the expert outputs during denoising. The final conditional estimate is computed as a weighted average:
\begin{equation}
    \tilde{e}(z_t, t, c) = \sum_{i=1}^N \hat{\Omega}^{t, \text{up}}_i e_{\theta_i}(z_t, t, c) \label{eq:weightedavg}
\end{equation}
We interpret cosine similarity in the noise prediction in the latent space as a proxy for LoRA confidence or relevance: patches where LoRA strongly deviates from the base model are assumed to reflect greater domain-specific influence. Because diffusion features strongly correlate with spatial layout, localised divergence indicates active, in-domain modification by the adapter. This follows from the low-rank structure of LoRA, as the update $\Delta W = AB$ has a negligible effect on inputs lying outside its learned subspace, which often correspond to out-of-distribution inputs. As such, high similarity indicates that the adapter is inactive, whereas lower similarity suggests in-distribution behaviour. A theoretical motivation for similarity-based weighting is included in~\cref{app:theory}.

\subsection{Re-centering guidance}\label{sec:recenter}
\begin{figure}[t]
    \centering
        \includegraphics[width=0.5\textwidth]{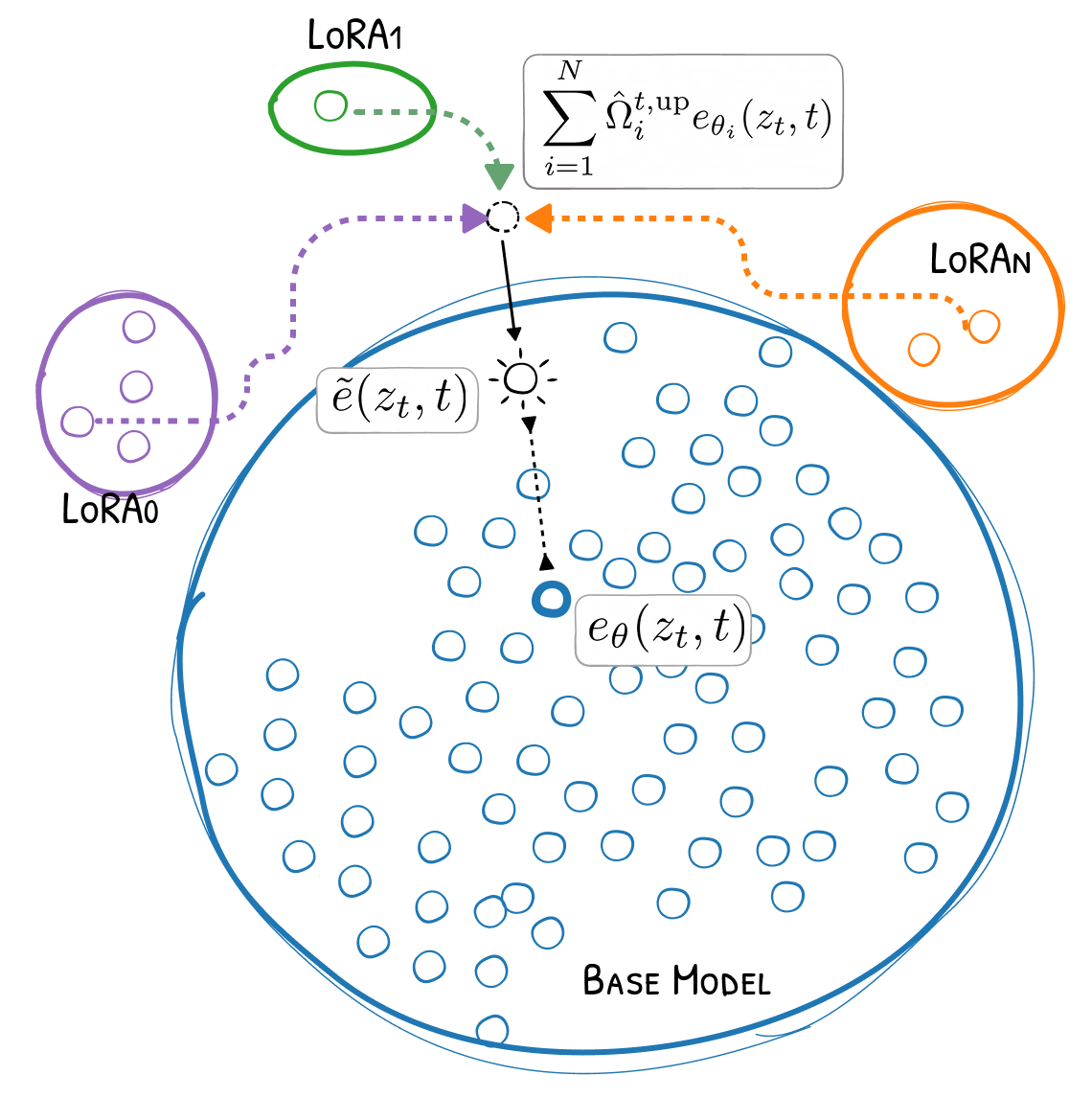}
        \caption{Visualisation of the effect of re-centering guidance (\eqref{eq:recenter}) on the unconditional noise score. Re-centering ensures the $p(x | c)$ is not over-emphasising implausible or under-trained regions of the collective data distribution after CFG. When the scores are similar, the transformation is not significant, but when there is a large adjustment, the difference is in the direction towards more probable samples.}
        \label{fig:method_recenter}
\end{figure}

To address Observation 1, we note that each LoRA $\theta_i$ implicitly encodes its own training distribution $p_{\text{LoRA}_i}(x)$, so its unconditional noise $e_{\theta_i}(z_t, t)$ diverges from the base $e_{\theta}(z_t, t)$ (\cref{fig:basemodelnoise}). Since CFG (\eqref{eq:cfg}) extrapolates between conditional and unconditional terms, the guidance term $e_{\theta_i}(z_t, t, c) - e_{\theta_i}(z_t, t)$ becomes a biased estimator of $\nabla_x \log p(x|c) - \nabla_x \log p(x)$, skewed by $p_{\text{LoRA}_i}(x)$. With multiple LoRAs active, these unconditional terms can additionally conflict, lowering subject fidelity under standard CFG.

To mitigate this drift, we propose ``re-centering'' the guidance computation by incorporating the unconditional base model output. Specifically, we use the average of the base model and LoRA-weighted unconditional outputs in CFG so that the final collective guidance $\hat{e}( z_t, t, c)$ is then calculated as follows:

\begin{equation}\label{eq:recenter}
    \begin{split}
        \Tilde{e}( z_t, t) &=  \lambda \sum_{i=1}^N \hat{\Omega}^{t, \text{up}}_i e_{\theta_i}( z_t, t) + (1 - \lambda) e_{\theta}( z_t, t) \\
        \hat{e}( z_t, t,c)  &= \Tilde{e}( z_t, t ) + s \left[ \Tilde{e}( z_t, t, c) - \Tilde{e}( z_t, t)\right] .
    \end{split}      
\end{equation}

We fix $\lambda=0.5$ across all experiments, informed by a qualitative ablation (\cref{app:ablation}). A visual representation of the re-centering method can be seen in~\cref{fig:method_recenter}.

\subsection{Dynamic module selection}
The Multi-LoRA composition task is defined under the assumption that only a known subset of LoRA adapters—those relevant to the current generation task—are loaded. This restricts flexibility, as it requires prior knowledge of which LoRAs are needed, and contradicts the goal of a truly inference-time, modular composition system. We propose expanding the task to a dynamic selection setting, where all available LoRA adapters are loaded into the model that dynamically selects which ones to activate based on the input. To address the dynamic setting, we propose using only the top-$k$ most distant LoRAs at each timestep $t$. We first perform a hard masking step by selecting the top-$k$ most relevant LoRA experts using a similarity-based gating metric $\Omega^t$. Specifically, we compute:
\begin{equation}
    \begin{split}
        \mathcal{I}_k &= \mathrm{TopK}(1 - \Omega^t, k) \\
        \tilde{\Omega}^t_i &=
        \begin{cases}
            \Omega^t_i & \text{if } i \in \mathcal{I}_k \\
            \infty & \text{otherwise}
        \end{cases} \\
        \hat{\Omega}^t_i &= \mathrm{softmin}_\tau(\tilde{\Omega}^t)
    \end{split}
\end{equation}
The $\hat{\Omega}^t_i$ is then upscaled and reshaped as described in~\cref{sec:loratorio}, so that it can be used in the subsequent weighted average and re-centering steps. For fair comparison against the static setting, we set $k=C$ where $C$ is the number of distinct concepts in the prompt; a fully zero-shot variant replacing TopK with a similarity threshold is left to future work. We include exploratory retrieval results for such a variant in~\cref{app:cost}.

\section{Experimental Results}\label{sec:experiments}
\subsection{Implementation Details}
For our experiments, we follow the setup of~\citep{zhongmulti2024}, using \textit{stable-diffusion-v1.5}~\citep{rombach_high-resolution_2022} as the backbone for all \textit{ComposLoRA} tests. We use the ``Realistic\_Vision\_V5.1'' and ``Counterfeit-V2.5'' checkpoints for realistic and anime-style images, respectively. For experiments with a Flux base model~\citep{flux2024}, we use the ``black-forest-labs/FLUX.1-dev'' checkpoint. For the realistic subset, we use 100 denoising steps, a guidance scale \( s = 7 \), and image size \( 1024 \times 768 \); for the anime subset, we use 200 steps, \( s = 10 \), and \( 512 \times 512 \) resolution. DPM-Solver++~\citep{lu2022dpm} is used as the sampler, with all LoRAs scaled by a weight of $0.8$. We empirically set an adaptive temperature $\tau = 1 / ((T-t) * 10)$, which decreases as denoising progresses, yielding softer gating at high-noise steps and sharper attribution as spatial structure emerges. The ablation in~\cref{tab:ablation} shows constant $\tau = 1$ shifts \clipscore by less than $0.25$ points, within seed variance, yet produces visibly inferior identity preservation at concept boundaries. For all experiments, we set the size of each patch to $2\times2$. Since our method operates at inference time, all experiments are run on a single RTX A6000 GPU. Results are averaged over three runs.

\subsection{\clipscore}
We employ \clipscore\citep{hessel2021clipscore} to evaluate how well the generated images match the text prompt, shown in~\cref{tab:clipscores}. Even though \clipscore does not evaluate compositional quality, acting more as a bag of words~\citep{zhongmulti2024}, it is still an important indicator of text-to-image fidelity. The multi-LoRA composition task becomes monotonically harder as $N$ increases: each additional adapter expands the space of inter-LoRA conflicts, increases the proportion of patches that lie out-of-distribution for at least one adapter, and amplifies cumulative domain drift in the unconditional score (\cref{fig:basemodelnoise}). This is reflected in the \clipscore profile of every prior method, all of which degrade from $N=2$ to $N=5$. LoRAtorio is the only method whose \clipscore does not collapse with $N$, surpassing the strongest prior method by up to 1.3\% at $N \geq 3$ and remaining competitive at $N=2$ where adversarial conflicts are minimal. We view robustness across $N$ rather than absolute performance at any single $N$ as the meaningful axis of evaluation. A breakdown by subset can be seen in~\cref{app:results}, and an ablation in~\cref{app:ablation}.

\begin{table*}[t]
\centering
\caption{\clipscore of LoRAtorio against previous composition methods on \textit{ComposLoRA}.}
\label{tab:clipscores}
\begin{tabular}{llllll}
\toprule

Model           & $N=2$           & $N=3$  & $N=4$            & $N=5$          & Avg.              \\
\midrule
Na\"{\i}ve~\citep{rombach_high-resolution_2022} & 35.014          & 34.927          & 34.384          & 33.809          & 34.534          \\
Merge~\citep{huggingface2024merge}         & 33.726          & 34.139          & 33.399          & 32.364          & 33.407          \\
Switch~\citep{zhongmulti2024}              & 35.394          & 35.107          & 34.478          & 33.475          & 34.614          \\
Composite~\citep{zhongmulti2024}           & 35.073          & 34.082          & 34.802          & 32.582          & 34.135          \\
LoraHub~\citep{huang2023lorahub}           & {\ul 35.681}    & 35.127          & 34.970          & 33.485          & 34.816          \\
Switch-A~\citep{zou_cached_2025}           & 35.451          & 35.383          & 34.877          & 33.366          & 34.769          \\
CMLoRA~\citep{zou_cached_2025}             & 35.422          &      35.215     & 35.208          & 34.341          & 35.047          \\
MultLFG~\citep{roy2025multlfg}            & \textbf{36.570} & {\ul 36.125}    & {\ul 36.180}    & {\ul 35.920}    & {\ul 36.199}    \\
LoRAtorio                                  & 35.236          & \textbf{36.426} & \textbf{37.136} & \textbf{36.626} & \textbf{36.356} \\ 
\bottomrule
\end{tabular}

\end{table*}

% \subsection{Qualitative Analysis}

\subsection{\gpt Evaluation}
To assess the compositional and aesthetic qualities of our method, we employ a~\gpt\nobreakdash-based evaluation as outlined in the \textit{ComposLoRA} testbed~\citep{zhongmulti2024}, scoring LoRAtorio against Switch, Composite, Merge and CMLoRA across two dimensions, ``Composition Quality'' and ``Image Quality''. MultLFG~\citep{roy2025multlfg} cannot be included in qualitative comparisons as neither code nor evaluation images have been released at the time of writing. Scores and Win Rates can be seen in~\cref{fig:gpt4vquality} and~\cref{fig:winrate}, respectively. LoRAtorio outperforms previous works both in terms of average scores and win rate, \ie pairwise comparison, closely followed by Switch. Additional results, including qualitative comparisons that corroborate this ranking, can be seen in~\cref{app:results}.

\begin{figure*}[t]
    \begin{subfigure}[t]{0.53\textwidth}
        \centering
        \begin{tikzpicture}
             \begin{axis}[
                ybar=1pt,
                bar width=15pt,
                width=\textwidth,
                height=6cm,
                ylabel={Score},
                ylabel shift=-5pt, 
                yticklabel style={inner sep=1pt},
                symbolic x coords={Composition Qual., Image Qual.},
                xtick=data,
                ymin=5,
                ymax=10,
                enlarge x limits=0.5,
                legend style={
                    at={(0.5,-0.15)},
                    anchor=north,
                    legend columns=3,
                    font=\footnotesize
                },
                nodes near coords,
                nodes near coords align=center,
                every node near coord/.append style={
                    font=\footnotesize,
                    anchor=south,
                    text=black
                },
                x label style={at={(axis description cs:0.5,-0.12)},anchor=north},
                axis lines*=left,
                tick style={draw=none},
                group style={
                    group size=2 by 1,
                    horizontal sep=0.07cm,
                    ylabels at=edge left
                },
                cycle list={
                    {fill=teagreen},      % LoRAtorio
                    {fill=pastel2},       % Merge
                    {fill=pastel3},       % Compose
                    {fill=pastelpink},    % Switch
                    {fill=pastelorange},  % CMLoRA
                }
            ]
            
            \addplot coordinates {(Composition Qual., 7.55) (Image Qual., 9.19)};
            \addlegendentry{LoRAtorio}
            
            \addplot coordinates {(Composition Qual., 7.04) (Image Qual., 8.82)};
            \addlegendentry{Merge}
            
            \addplot coordinates {(Composition Qual., 6.72) (Image Qual., 8.99)};
            \addlegendentry{Composite}
            
            \addplot coordinates {(Composition Qual., 7.22) (Image Qual., 9.13)};
            \addlegendentry{Switch}
            
            \addplot coordinates {(Composition Qual., 5.70) (Image Qual., 8.44)};
            \addlegendentry{CMLoRA}
            
            \end{axis}
            \end{tikzpicture}
    \caption{Composition and Image Quality comparison.}
    \label{fig:gpt4vquality}
    \end{subfigure}
    \hfill
    \begin{subfigure}[t]{0.47\textwidth}
        \centering
        \begin{tikzpicture}
        \begin{axis}[
            ybar stacked,
            width=\textwidth,
            height=6cm,
            bar width=30pt,
            enlarge x limits=0.2,
            ylabel={Percentage},
            ylabel shift=-5pt, 
            yticklabel style={inner sep=1pt},
            symbolic x coords={Switch, Composite, Merge, CMLoRA},
            xtick=data,
            ymin=0,
            ymax=100,
            ytick={0,20,40,60,80,100},
            xticklabel style={font=\footnotesize},
             legend style={
                at={(0.5,-0.35)},
                anchor=south,
                legend columns=-1,
                font=\footnotesize
            },
            nodes near coords,
            every node near coord/.append style={font=\small, color=black},
            major grid style={gray!30},
            grid=major,
        ]
        
        \legend{Win, Tie, Lose}
        \addplot+[
            ybar,
            pattern={north east lines},
            pattern color=teagreen,
            draw=gray
        ] coordinates {
            (Switch,48.72) 
            (Composite,58.97) 
            (Merge,56.41) 
            (CMLoRA,76.92) 
        };
        
        % Tie - pastel yellow (0s, still included for completeness)
        \addplot+[
            ybar,
            pattern=north east lines,
            pattern color=pastelorange,
            draw=gray
        ] coordinates {
            (Switch,12.82) 
            (Composite,15.38) 
            (Merge,12.82) 
            (CMLoRA,7.69) 
        };
        \addplot+[
            ybar,
            pattern=north east lines,
            pattern color=pastelpink,
            draw=gray
        ] coordinates {
            (Switch,38.46) 
            (Composite,25.64) 
            (Merge,30.77) 
            (CMLoRA,15.38) 
        };
        \end{axis}
        \end{tikzpicture}
        \caption{Overall win rate of LoRAtorio.}
        \label{fig:winrate}
    \end{subfigure}
    \caption{\gpt Evaluation on \textit{ComposLoRA} against previous SoTA works.}
\end{figure*}

\subsection{Human Evaluation}
Further to the~\gpt evaluation, we employ human experts to assess LoRAtorio qualitatively against previous works, as described by~\citep{zou_cached_2025} across four criteria: Element Integration, Spatial Consistency, Semantic Accuracy, and Aesthetic Quality. As with the \gpt evaluation, MultLFG is omitted because no source code or evaluation images are publicly available.
The results shown in~\cref{tab:human} corroborate the~\gpt evaluation, with LoRAtorio outperforming all previous works closely followed by Switch. Details of the interface and definitions for the human evaluation can be seen in~\cref{app:human}.

\begin{table}[h]
\caption{Human Evaluation of our Method against previous SoTA along four qualitative axes.}
\label{tab:human}
\centering
\resizebox{0.9\textwidth}{!}{%
\begin{tabular}{lcccc}
\toprule
          & \multicolumn{1}{l}{Element Integration} & \multicolumn{1}{l}{Spatial Consistency} & \multicolumn{1}{l}{Semantic Accuracy} & \multicolumn{1}{l}{Aesthetic Quality} \\
\midrule
LoRAtorio & \textbf{7.64}                           & \textbf{7.58}                           & \textbf{7.33}                         & \textbf{6.83}                         \\
CMLoRA      & 5.63                                    & 5.58                                    & 6.08                                  & 5.25                                  \\
Composite   & 6.46                                    & 6.71                                    & 6.71                                  & 6.46                                  \\
Switch      & {\ul 7.57}                              & {\ul 7.50}                              & {\ul 6.88}                            & {\ul 6.71}                            \\
Merge       & 6.83                                    & 6.71                                    & 6.58                                  & 6.08   \\
\bottomrule
\end{tabular}
}
\end{table}

\subsection{Dynamic Module Selection}
When all LoRAs are loaded simultaneously, LoRA Merge collapses to incoherent outputs (\cref{tab:clipopen} and~\cref{app:qual}): even sparse, low-norm LoRA weights interfere meaningfully with denoising under out-of-distribution conditioning, motivating reliable per-step selection. LoRAtorio maintains high \clipscore in this setting, with minimal influence from unrelated adapters. We provide a comparison of the computational cost (MACs and latency) against previous works in~\cref{app:cost}.

\begin{table}[ht]
\caption{\clipscore of LoRAtorio against previous SoTA on \textit{ComposLoRA} in a dynamic module selection setting, where $N$ is the number of LoRA experts needed.}\label{tab:clipopen}
\centering
\resizebox{0.7\textwidth}{!}{%
\begin{tabular}{llllll}
\toprule
                    & $N=2$  & $N=3$  & $N=4$  & $N=5$  & Avg. \\
\midrule
LoRAtorio           & \textbf{34.593} & \textbf{35.563} & \textbf{36.480} & \textbf{37.028} & \textbf{35.916} \\
Na\"{\i}ve          & 35.014          & 34.927 & 34.384 & 33.809 & 34.534 \\
Merge               & 27.167          & 27.151 & 27.023 & 27.272 & 27.153 \\

\bottomrule
\end{tabular}
}
\end{table}

\subsection{Extension to Rectified-Flow Architectures}

Our method is architecturally agnostic: any latent diffusion model whose denoiser produces a structured noise prediction is compatible with patch-level similarity gating. To illustrate this generality, we extend LoRAtorio to a rectified-flow backbone~\citep{liu2023flow}, FLUX.1-dev. As FLUX is transformer-based and its latent space is natively tokenised, we apply the gating directly on the token outputs and omit explicit patch tokenisation; FLUX also embeds guidance signals inside cross-attention rather than via the standard CFG construction, so the re-centering step (which targets CFG drift) is structurally inapplicable and is omitted (\cref{app:fluxlora}). The FLUX baselines available out-of-the-box are limited, as adapting prior multi-LoRA methods to rectified-flow requires non-trivial re-engineering. The purpose of this evaluation is therefore to demonstrate the principle's portability rather than to produce a head-to-head benchmark; \cref{tab:fluxclip} confirms that the trends from SD1.5 transfer cleanly to a stronger, fundamentally different backbone.

\begin{table}[htbp]
    \centering
    \caption{\clipscore of LoRAtorio against selected composition methods, using Flux architecture.}\label{tab:fluxclip}
    
    \begin{subtable}[t]{0.49\textwidth}
        \centering
        \caption{Static Modules}
        \resizebox{\textwidth}{!}{%
        \begin{tabular}{llllll}
            \toprule
            Model           & $N=2$           & $N=3$           & $N=4$                        & $N=5$        &  Avg.         \\
            \midrule
            Na\"{\i}ve      & 33.125          & 34.999          & {\ul{37.048}}              & {\ul{38.568}}  &   {\ul 35.935}    \\
            Merge           & {\ul{33.733}}   & {\ul{35.134}}   & {35.830}                   & {36.590}       &   35.322   \\
            LoRAtorio       & \textbf{33.992} & \textbf{36.033} & \textbf{37.781}            & \textbf{39.368}&   \textbf{36.794}   \\
            \bottomrule
            \end{tabular}
            }
    \end{subtable}
    \hfill
    \begin{subtable}[t]{0.49\textwidth}
        \centering
        \caption{Dynamic Module Selection}
        \resizebox{\textwidth}{!}{%
        \begin{tabular}{llllll}
            \toprule
                       & $N=2$  & $N=3$  & $N=4$         & $N=5$         & Avg. \\
            \midrule
            Na\"{\i}ve & {\ul 33.125} & {\ul 34.999} & {\ul{37.048}} & {\ul{38.568}} &   {\ul{35.935}}    \\
            Merge      & 25.850 & 27.858 & 29.726  & 30.997 & 28.608 \\
            LoRAtorio  & \textbf{33.284} & \textbf{35.753} & \textbf{37.910} & \textbf{38.861} & \textbf{36.452} \\
            \bottomrule
            \end{tabular}
            }
    \end{subtable}
\end{table}

\section{Related Work}

\subsection{Text-to-image Generation}
Composable image generation aims to faithfully integrate multiple user-specified concepts. Early methods relied on layout or scene-graph conditioning~\citep{johnson2018image, song2021scorebased, gafni2022make}, while recent works modify the diffusion generative process for multi-concept prompts~\citep{feng2023trainingfree, huang2023composer, kumari2023multi, lin2023designbench, ouyang2025k}. However, these approaches often require prompt engineering or architectural changes and struggle with rare, user-defined elements. Alternatively, composing independently trained modules~\citep{du2020compositional, liu2021learning, li2023composing, simsar2025loraclr} requires extensive fine-tuning and scales poorly. In contrast, our train-free, instance-level framework leverages LoRA adapters for fine-grained, spatially-aware concept integration.

\subsection{LoRA-Based Skill Composition}
Low-Rank Adaptation (LoRA) enables efficient personalization in diffusion models~\citep{ruiz2023dreambooth, sohn2023styledrop}. Strategies for composing multiple LoRAs broadly fall into two categories:

\textbf{Trainable Approaches:} LoRAHub~\citep{huang2023lorahub} and ZipLoRA~\citep{shah2024ziplora} use few-shot demonstrations to learn a linear combination of LoRA weights. Others propose mixture-of-experts (MoE)~\citep{zhu_mole_2024} or hypernetworks~\citep{shenaj_lorarar_2024, ruiz2024hyperdreambooth} to generate target weights. These methods demand additional training data and struggle in zero-shot or open-vocabulary settings.

\textbf{Train-free Approaches:} These methods manipulate weights or inference directly. LoRA Merge~\citep{huggingface2024merge} uses weight-level arithmetic, while CLoRA~\citep{meral2024clora} targets attention maps. LoRA Switch and Composite~\citep{zhongmulti2024} alternate or aggregate outputs during denoising, and~\cite{zou_cached_2025} adds frequency-based scheduling and caching. MultLFG~\citep{roy2025multlfg} employs frequency-domain guidance but necessitates costly per-step RGB decoding, bypassing the more intuitive latent noise space. As the number of LoRAs increases, these methods often suffer from semantic conflicts and instability. Crucially, they ignore interactions with the base model, overlook domain shifts caused by fine-tuning, and are limited to text conditions.

Inspired by spatial composition techniques like CutMix~\citep{yun2019cutmix} and token-level fusion~\citep{wang2024tokencompose}, LoRAtorio applies these principles in the latent space. While resembling an MoE~\citep{moe}, our method is entirely train-free: it uses intrinsic cosine similarity between LoRA and base model latents for zero-shot gating, operating directly within the semantic latent space rather than relying on learned routing or image-space features.

\section{Conclusion}
In this paper, we present a novel, train-free approach to multi-LoRA composition through the introduction of LoRAtorio, a method grounded in intrinsic model behaviour. Motivated by empirical observations of domain drift and latent-space divergence, our method leverages spatially-aware similarity to dynamically weight LoRA contributions at the patch level. We further propose a modification to classifier-free guidance that incorporates the base model’s unconditional signal, improving robustness in out-of-domain scenarios. Extending beyond static composition, we formulate the task as one of dynamic module selection, enabling inference-time adaptability in settings where irrelevant skills are loaded to the base model. Our approach achieves state-of-the-art performance on \textit{ComposLoRA} with \clipscore that does not deteriorate as the number of composed adapters grows, and we further demonstrate the portability of the intrinsic-similarity principle to a rectified-flow backbone.

%%%%%%%%%%%%%%%%%%%%%%%%%%%%%%%%%%%%%%%%%%%%%%%%%%%%%%%%%%%%

\clearpage
\appendix
% Make \cref to sections defined here render as ``Appendix X'' instead of ``Section X''.
\crefalias{section}{appendix}
\crefalias{subsection}{appendix}

\section{Ethics Statement}

This work examines the capabilities of generative AI models, including those enhanced with community-provided LoRAs. While generative tools offer valuable opportunities for creative and technical innovation, they also carry significant risks, including misuse for deceptive content, reinforcement of harmful biases, and uncertainty around authorship and licensing.

 We do not condone the misuse of generative models, including for misinformation, harassment, or any activity that infringes on the rights of others. This work is not licensed or intended for commercial or for-profit use. We encourage future users and researchers to carefully consider the ethical and legal implications of models or data.

\section{Theoretical Motivation for Similarity-Based Weighting.}\label{app:theory}
LoRA introduces a low-rank update to a weight matrix $W \in \mathbb{R}^{d \times k}$ in the form $\Delta W = AB$, where $A \in \mathbb{R}^{d \times r}$, $B \in \mathbb{R}^{r \times k}$, and $ r \ll \min(d, k) $~\citep{hu2022lora}. This constrains the update to lie in a low-dimensional subspace of the weight space, limiting the directions in which the model can adapt. Such low-rank adaptation has been shown to improve parameter efficiency and mitigate catastrophic forgetting~\citep{biderman2024lora}.

More precisely, the LoRA update acts on inputs $x \in \mathbb{R}^k$ by first projecting via $A$, $A x \in \mathbb{R}^r$, then mapping back to output space via $B$. The effective input subspace to which the adapter responds is the row space of $A$, \ie, inputs $x$ for which $Ax \neq 0$. For inputs $x'$ approximately orthogonal to this subspace, $A x' \approx 0$ and thus
\begin{equation}
    \Delta W x' = (A x') B \approx 0,
\end{equation}
implying
\begin{equation}
    W x' + \Delta W x' \approx W x'.
\end{equation}
Hence, the LoRA adapter has a negligible effect on inputs lying outside its learned subspace, which often correspond to out-of-distribution (OOD) inputs, and subsequent non-linearities in deep learning models further mitigate the effect of LoRAs.
Consequently, the latent outputs of the LoRA-augmented model and the base model are similar for OOD inputs. This motivates using the cosine similarity between their latent outputs as a proxy for the adapter’s confidence or relevance: high similarity indicates that the adapter is inactive or uncertain (OOD), whereas lower similarity suggests in-distribution behaviour where the adapter actively modifies the model output. This observation underpins our use of cosine similarity in LoRAtorio.

\subsection{Intuition Behind Similarity-Based Weighting}
\label{sec:intuition_weighting}

To supplement the theoretical motivation above, we provide an intuitive explanation for our patch-level cosine similarity mechanism. Prior works have established that diffusion features correlate strongly with the spatial layout of the generated image. For a fixed initial noise seed, different LoRA adapters maintain a consistent ``structural blueprint'' that closely mirrors the base model, particularly when the text condition is out-of-distribution for the adapter. 

LoRAtorio leverages this spatial consistency. When a specific LoRA is highly relevant to a given spatial patch (e.g., a character's face), it actively modifies the local latent representation, causing a significant, meaningful divergence from the base model's prediction. The SoftMin function subsequently assigns a higher weight to these divergent (low-similarity) patches. This localized gating mechanism prevents the ``blurred average'' phenomenon observed in simple weight-merging techniques by ensuring that only the most spatially relevant experts contribute to specific regions of the final image.

\section{Preliminary Motivational Experiments}
\label{sec:kl_divergence}

To further substantiate the visual intuition presented in~\cref{fig:basemodelnoise}, and to quantitatively support Observation 1 and Observation 2, we perform an empirical analysis of the latent space distributions. While visualisations such as t-SNE provide an intuitive understanding of the domain shift caused by Low-Rank Adaptation (LoRA), we supplement this evaluation with quantitative metrics.

We isolate the unconditional noise predictions from the base model and the LoRA-adapted models to measure the extent to which specialised LoRAs diverge from the base distribution. For this experiment, we extract the unconditional latent noise predictions $e_{\theta}(z_0, 0)$ and $e_{\theta_i}(z_0, 0)$ across $100$ random seeds. To make the comparison computationally tractable while preserving the core structural variances, we apply Principal Component Analysis (PCA) to project the high-dimensional latents into a 2D space that retains 95\% of the variance. 

Over these estimated 2D density distributions, we calculate the Kullback-Leibler (KL) divergence from the base model distribution $P$ to the $i^{th}$ LoRA distribution $Q_i$:

\begin{equation}
    D_{KL}(P \parallel Q_i) = \int P(x) \log\left(\frac{P(x)}{Q_i(x)}\right) dx
\end{equation}

As shown in Table~\ref{tab:kl_divergence}, highly specialised domains—particularly characters and specific artistic styles—exhibit a significant KL divergence from the base model. Backgrounds and generic objects, which are inherently closer to the diverse pre-training data of the base model, show comparatively lower divergence. 

\begin{table}[ht]
    \centering
    \caption{KL Divergence of single LoRA adapters against the base model.}
    \label{tab:kl_divergence}
    \begin{tabular}{lc}
        \toprule
        \textbf{LoRA Model} & \textbf{KL Divergence}  \\
        \midrule
        Character 1 & 34.81 \\
        Character 2 & 35.09 \\
        Character 3 & 36.15 \\
        Style 1 & 36.44 \\
        Style 2 & 36.31 \\
        Object 1 & 35.52 \\
        Object 2 & 21.49 \\
        Background 1 & 21.34 \\
        Background 2 & 9.43 \\
        \bottomrule
    \end{tabular}
\end{table}

\section{Experimental Results}
\label{app:results}

\begin{table}[htbp]
    \centering
    \caption{ClipScores for \textit{ComposLoRA} on anime and reality subsets.}
    \label{tab:combinedclip}

    \begin{minipage}{0.48\textwidth}
        \centering
        \begin{subtable}[t]{\textwidth}
            \centering
            \caption{Anime – Static Modules}
            \resizebox{\textwidth}{!}{
                \begin{tabular}{llllll}
                \toprule
                Model       & $N=2$           & $N=3$           & $N=4$           & $N=5$           & Avg.            \\ 
                \midrule
                Merge       & 35.136          & 35.421          & 34.164          & 32.636          & 34.339          \\
                Switch      & 35.285          & 35.482          & 34.532          & 34.148          & 34.861          \\
                Composite   & 34.343          & 34.378          & 34.161          & 32.936          & 33.955          \\
                LoraHub     & 35.316          & 35.525          & 34.476          & 33.885          & 34.801          \\
                Switch-A    & 35.705          & 35.912          & 35.661          & 34.479          & 35.439          \\
                CMLoRA      & 35.556          & 35.555          & 35.791          & 35.691          & 35.648         \\
                MultLFG     & \textbf{36.720} & {\ul 36.130}    & {\ul 36.450}    & {\ul 36.220}    & {\ul 36.380}    \\
                LoRAtorio   & {\ul 36.156}  & \textbf{36.930}   & \textbf{36.864} & \textbf{36.162} & \textbf{36.528} \\
                \bottomrule
                \end{tabular}
            }
        \end{subtable}
    \end{minipage}
    \hfill
    \begin{minipage}{0.48\textwidth}
        \centering
        \begin{subtable}[t]{\textwidth}
            \centering
            \caption{Reality – Static Modules}
            \resizebox{\textwidth}{!}{
            \begin{tabular}{llllll}
                \toprule
                
                Model      & $N=2$           & $N=3$           & $N=4$           & $N=5$           & Avg.            \\ 
                \midrule
                Merge      & 32.316          & 32.857          & 32.633          & 32.091          & 32.474          \\
                Switch     & 35.502          & 34.731          & 34.424          & 32.801          & 34.365          \\
                Composite  & 35.804          & 33.786          & 35.443          & 32.228          & 34.315          \\
                LoraHub    & {\ul 36.045}    & 34.729          & 35.463          & 33.084         & 35.412          \\
                Switch-A   & 35.196          & 34.854          & 34.694          & 32.252          & 34.249          \\
                CMLoRA     & 35.559          & 35.842          & 34.501          & 33.588          & 34.873          \\
                MultLFG    & \textbf{36.420} & {\ul 36.120}    & {\ul 35.910}    & {\ul 35.620}    & {\ul 36.018}    \\
                LoRAtorio  & 34.316          & \textbf{35.922} & \textbf{37.408} & \textbf{37.090} & \textbf{36.184} \\
                \bottomrule
            \end{tabular}
            }
        \end{subtable}
    \end{minipage}

    \vspace{1em}

    \begin{minipage}{0.48\textwidth}
        \centering
        \begin{subtable}[t]{\textwidth}
            \centering
            \caption{Anime – Dynamic Modules}
            \resizebox{\textwidth}{!}{
                \begin{tabular}{llllll}
                    \toprule
                                        & $N=2$  & $N=3$  & $N=4$  & $N=5$  & Avg. \\
                    \midrule
                    LoRAtorio           & \textbf{35.328} & \textbf{35.931} & \textbf{36.332} & \textbf{36.184} & \textbf{35.944}\\
                    Na\"{\i}ve          & 35.014 & 34.927 & 34.384 & 33.809 & 34.534\\
                    Merge               & 30.953 & 30.767 & 30.352 & 30.488 & 30.640\\            
                    \bottomrule
                \end{tabular}
            }
        \end{subtable}
    \end{minipage}
    \hfill
    \begin{minipage}{0.48\textwidth}
        \centering
        \begin{subtable}[t]{\textwidth}
            \centering
            \caption{Reality – Dynamic Modules}
            \resizebox{\textwidth}{!}{
                \begin{tabular}{llllll}
                    \toprule
                                        & $N=2$  & $N=3$  & $N=4$  & $N=5$  & Avg. \\
                    \midrule
                    LoRAtorio           & 33.858 & \textbf{35.194} & \textbf{36.627} & \textbf{37.871} & \textbf{35.888}\\
                    Na\"{\i}ve          & \textbf{35.014} & 34.927 & 34.384 & 33.809 & 34.534\\
                    Merge               & 23.381 & 23.534 & 23.693 & 24.055 & 23.666\\            
                    \bottomrule
                \end{tabular}
            }
        \end{subtable}
    \end{minipage}
\end{table}
Further to the main experimental results in~\cref{sec:experiments}, we show \clipscore for the subsets of \textit{ComposLoRA} in~\cref{tab:combinedclip} for anime and reality subsets in both static and dynamic module settings. LoRAtorio maintains strong performance in both subsets, having the highest weighted average score compared to previous SoTA. This is consistent for both static and dynamic module selection. We observe similar trends in performance within each subset for the number of LoRAs included, with LoRAtorio clearly outperforming other works on average. As expected, we also see weight merge collapsing in the dynamic module setting.

\begin{figure*}[htbp]
    \centering
     \begin{subfigure}[t]{0.49\textwidth}
        \includegraphics[width=\textwidth]{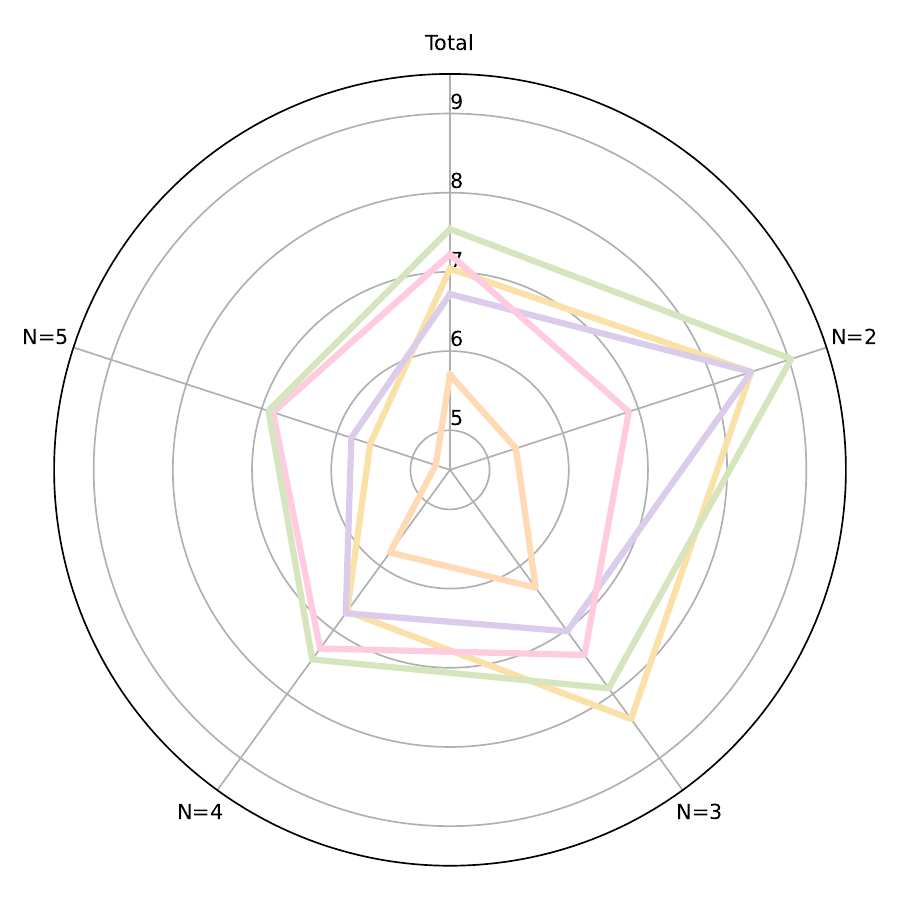}
        \caption{Composition Quality by $N$}
     \end{subfigure}
     \hfill
     \begin{subfigure}[t]{0.49\textwidth}
         \includegraphics[width=\textwidth]{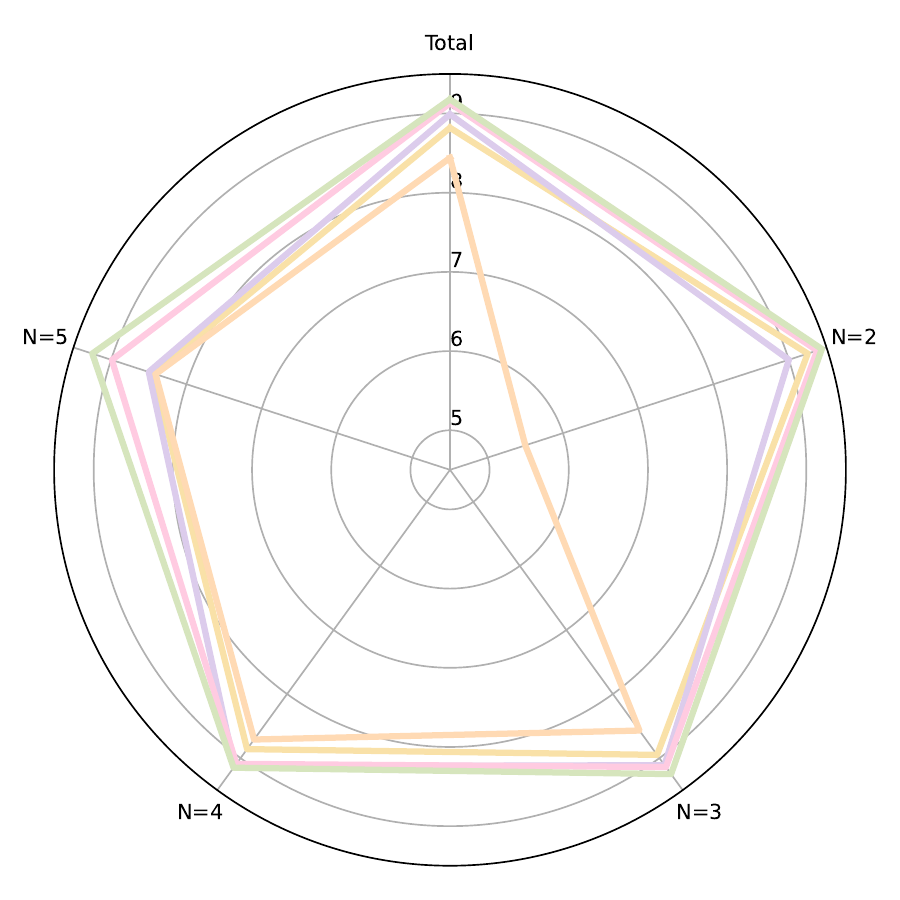}
         \caption{Image Quality by $N$}
     \end{subfigure}
     
     \begin{subfigure}[t]{\textwidth}
         \begin{tikzpicture}
            % --- Anime Chart ---
            \begin{axis}[
                ybar=1pt,
                bar width=12pt,
                width=0.5\textwidth,  % <-- Increased width
                height=6cm,
                ylabel={Score},
                ylabel shift=-5pt, 
                yticklabel style={inner sep=1pt},
                symbolic x coords={Composition Qual., , Image Qual.},
                xtick=data,
                ymin=5,
                ymax=10,
                enlarge x limits=0.5, % <-- Reduced empty space on the sides
                title={Anime},
                nodes near coords,
                nodes near coords align={vertical},
                every node near coord/.append style={
                    font=\footnotesize,
                    anchor=south,
                    text=black
                },
                cycle list={
                    {fill=teagreen},
                    {fill=pastel2},
                    {fill=pastel3},
                    {fill=pastelpink},
                    {fill=pastelorange},
                }
            ]
            
            \addplot coordinates {(Composition Qual., 7.4625) (,0) (Image Qual., 9.175)};
            \addplot coordinates {(Composition Qual., 6.8625) (,0) (Image Qual., 8.775)};
            \addplot coordinates {(Composition Qual., 6.725) (,0) (Image Qual., 8.9125)};
            \addplot coordinates {(Composition Qual., 7.37098) (,0) (Image Qual., 8.99464)};
            \addplot coordinates {(Composition Qual., 6.2375) (,0) (Image Qual., 8.5625)};
            
            \end{axis}
            
            % --- Reality Chart ---
            \begin{axis}[
                at={(9cm,0)}, % <-- Pushed slightly right to accommodate wider charts
                anchor=origin,
                ybar=1pt,
                bar width=12pt,
                width=0.5\textwidth, % <-- Increased width
                height=6cm,
                % ylabel={Score},
                % ylabel shift=-5pt,
                % yticklabel style={inner sep=1pt},
                symbolic x coords={Composition Qual., , Image Qual.},
                xtick=data,
                ymin=5,
                ymax=10,
                enlarge x limits=0.5, % <-- Reduced empty space on the sides
                title={Reality},
                nodes near coords,
                nodes near coords align={vertical},
                every node near coord/.append style={
                    font=\footnotesize,
                    anchor=south,
                    text=black
                },
                cycle list={
                    {fill=teagreen},
                    {fill=pastel2},
                    {fill=pastel3},
                    {fill=pastelpink},
                    {fill=pastelorange},
                }
            ]
            
            \addplot+[fill=teagreen] coordinates {(Composition Qual., 7.72368) (,0) (Image Qual., 9.27303)};
            \addplot+[fill=pastel2] coordinates {(Composition Qual., 7.22368) (,0) (Image Qual., 8.86842)};
            \addplot+[fill=pastel3] coordinates {(Composition Qual., 6.71053) (,0) (Image Qual., 9.06579)};
            \addplot+[fill=pastelpink] coordinates {(Composition Qual., 6.96053) (,0) (Image Qual., 9.18421)};
            \addplot+[fill=pastelorange] coordinates {(Composition Qual., 5.13158) (,0) (Image Qual., 8.31579)};
            
        \end{axis}
        
        \end{tikzpicture}
        \caption{Breakdown by Image Style}
     \end{subfigure}
     
        \begin{tikzpicture}
                \begin{axis}[
                    hide axis,
                    xmin=0, xmax=1,
                    ymin=0, ymax=1,
                    legend style={
                        at={(0.5,-0.1)},
                        anchor=north,
                        legend columns=3,
                        /tikz/every even column/.append style={column sep=0.5cm},
                        font=\small
                    },
                    cycle list={
                        {teagreen},      % LoRAtorio
                        {pastel2},       % Merge
                        {pastel3},       % Compose
                        {pastelpink},    % Switch
                        {pastelorange},  % CMLoRA
                    }
                ]
                \addlegendimage{area legend, fill=teagreen}
                \addlegendentry{LoRAtorio}
                \addlegendimage{area legend, fill=pastel2}
                \addlegendentry{Merge}
                \addlegendimage{area legend, fill=pastel3}
                \addlegendentry{Compose}
                \addlegendimage{area legend, fill=pastelpink}
                \addlegendentry{Switch}
                \addlegendimage{area legend, fill=pastelorange}
                \addlegendentry{CMLoRA}
                \end{axis}
            \end{tikzpicture}
        \caption{Composition and Image Quality scores of LoRAtorio against previous SoTA on \textit{ComposeLoRA}, by number of LoRA's included (a-b) and image style (c).}
        \label{fig:gpt4vqualitybyN}
\end{figure*}

\begin{figure*}[htbp]
    \centering
    
    % === SUBFIGURE: N=2 ===
    \begin{subfigure}[t]{0.48\textwidth}
    \centering
    \begin{tikzpicture}
    \begin{axis}[
        ybar stacked,
        bar width=20pt,
        ymin=0, ymax=100,
        symbolic x coords={Switch,Comp.,Merge,CMLoRA},
        xtick=data,
        xticklabel style={font=\tiny},
        ytick={0,50,100},
        yticklabels={},
        axis line style={gray},
        % tick style={gray},
        % tick label style={gray},
        nodes near coords,
        every node near coord/.append style={font=\tiny, color=black, anchor=center},
        title={\tiny $N = 2$},
        width=\textwidth,
        height=4.5cm,
        enlarge x limits=0.15
    ]
    \addplot+[pattern={north east lines}, pattern color=pastel_green, draw=gray] coordinates {
        (Switch,34) (Comp.,50) (Merge,50) (CMLoRA,50)
    };
    \addplot+[pattern={north east lines}, pattern color=pastel_yellow, draw=gray] coordinates {
        (Switch,33) (Comp.,25) (Merge,25) (CMLoRA,25)
    };
    \addplot+[pattern={north east lines}, pattern color=pastel_red, draw=gray] coordinates {
        (Switch,33) (Comp.,25) (Merge,25) (CMLoRA,25)
    };
    \end{axis}
    \end{tikzpicture}
    \end{subfigure}
    %
    % === SUBFIGURE: N=3 ===
    \begin{subfigure}[t]{0.48\textwidth}
    \centering
    \begin{tikzpicture}
    \begin{axis}[
        ybar stacked,
        bar width=20pt,
        ymin=0, ymax=100,
        symbolic x coords={Switch,Comp.,Merge,CMLoRA},
        xtick=data,
        xticklabel style={font=\tiny},
        ytick={0,50,100},
        yticklabels={},
        axis line style={gray},
        % tick style={gray},
        % tick label style={gray},
        nodes near coords,
        every node near coord/.append style={font=\tiny, color=black, anchor=center},
        title={\tiny $N = 3$},
        width=\textwidth,
        height=4.5cm,
        enlarge x limits=0.15
    ]
    \addplot+[pattern=north east lines, pattern color=pastel_green, draw=gray] coordinates {
        (Switch,43) (Comp.,57) (Merge,50) (CMLoRA,67)
    };
    \addplot+[pattern={north east lines},pattern color=pastel_yellow, draw=gray] coordinates {
        (Switch,14) (Comp.,14) (Merge,13) (CMLoRA,17)
    };
    \addplot+[pattern={north east lines}, pattern color=pastel_red, draw=gray] coordinates {
        (Switch,43) (Comp.,29) (Merge,38) (CMLoRA,17)
    };
    \end{axis}
    \end{tikzpicture}
    \end{subfigure}
    %
    
    % === SUBFIGURE: N=4 ===
    \begin{subfigure}[t]{0.48\textwidth}
    \centering
    \begin{tikzpicture}
    \begin{axis}[
        ybar stacked,
        bar width=20pt,
        ymin=0, ymax=100,
        symbolic x coords={Switch,Comp.,Merge,CMLoRA},
        xtick=data,
        xticklabel style={font=\tiny},
        ytick={0,50,100},
        yticklabels={},
        axis line style={gray},
        % tick style={gray},
        % tick label style={gray},
        nodes near coords,
        every node near coord/.append style={font=\tiny, color=black, anchor=center},
        title={\tiny $N = 4$},
        width=\textwidth,
        height=4.5cm,
        enlarge x limits=0.15
    ]
    \addplot+[pattern=north east lines, pattern color=pastel_green, draw=gray] coordinates {
        (Switch,53) (Comp.,71) (Merge,65) (CMLoRA,85)
    };
    \addplot+[pattern={north east lines}, pattern color=pastel_yellow, draw=gray] coordinates {
        (Switch,5) (Comp.,6) (Merge,6) (CMLoRA,8)
    };
    \addplot+[pattern={north east lines}, pattern color=pastel_red, draw=gray] coordinates {
        (Switch,42) (Comp.,24) (Merge,29) (CMLoRA,8)
    };
    \end{axis}
    \end{tikzpicture}
    \end{subfigure}
    %
    % === SUBFIGURE: N=5 ===
    \begin{subfigure}[t]{0.48\textwidth}
    \centering
    \begin{tikzpicture}
    \begin{axis}[
        ybar stacked,
        bar width=20pt,
        ymin=0, ymax=100,
        symbolic x coords={Switch,Comp.,Merge,CMLoRA},
        xtick=data,
        xticklabel style={font=\tiny},
        ytick={0,50,100},
        yticklabels={},
        axis line style={gray},
        % tick style={gray},
        % tick label style={gray},
        nodes near coords,
        every node near coord/.append style={font=\tiny, color=black, anchor=center},
        title={\tiny $N = 5$},
        width=\textwidth,
        height=4.5cm,
        enlarge x limits=0.15
    ]
    \addplot+[pattern=north east lines, pattern color=pastel_green, draw=gray] coordinates {
        (Switch,50) (Comp.,42) (Merge,50) (CMLoRA,50)
    };
    \addplot+[pattern=north east lines, pattern color=pastel_yellow, draw=gray] coordinates {
        (Switch,20) (Comp.,25) (Merge,20) (CMLoRA,20)
    };
    \addplot+[pattern=north east lines, pattern color=pastel_red, draw=gray] coordinates {
        (Switch,30) (Comp.,33) (Merge,30) (CMLoRA,30)
    };
    \end{axis}
    \end{tikzpicture}
    \end{subfigure}
    
    \vspace{0.5em}
    \caption{Win/Tie/Loss for LoRAtorio compared to previous SoTA across different number of LoRAs $\left( N \right)$ .}
    \label{fig:winbyN}
\end{figure*}
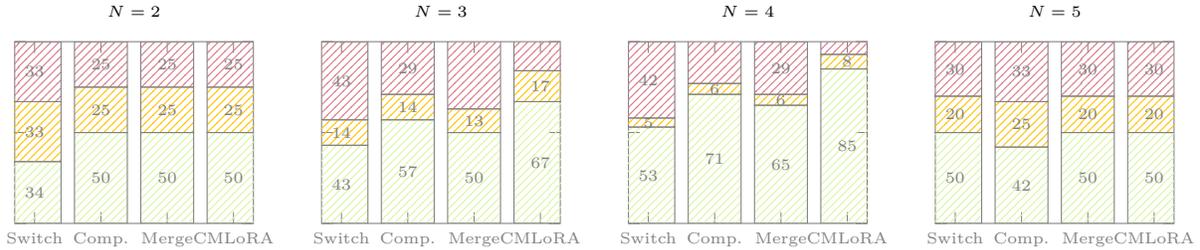

Further to the ClipScores, we present the~\gpt evaluation results by number of LoRAs included and by sub-set in~\ref{fig:gpt4vqualitybyN}. LoRAtorio maintains robust performance in all scenarios, showing strong composition and image quality. Finally, we include the win rate of our method by number of LoRAs included, showing SoTA performance, particularly as $N$ increases in~\cref{fig:winbyN}.

\begin{figure*}[htbp]
    \centering
    \includegraphics{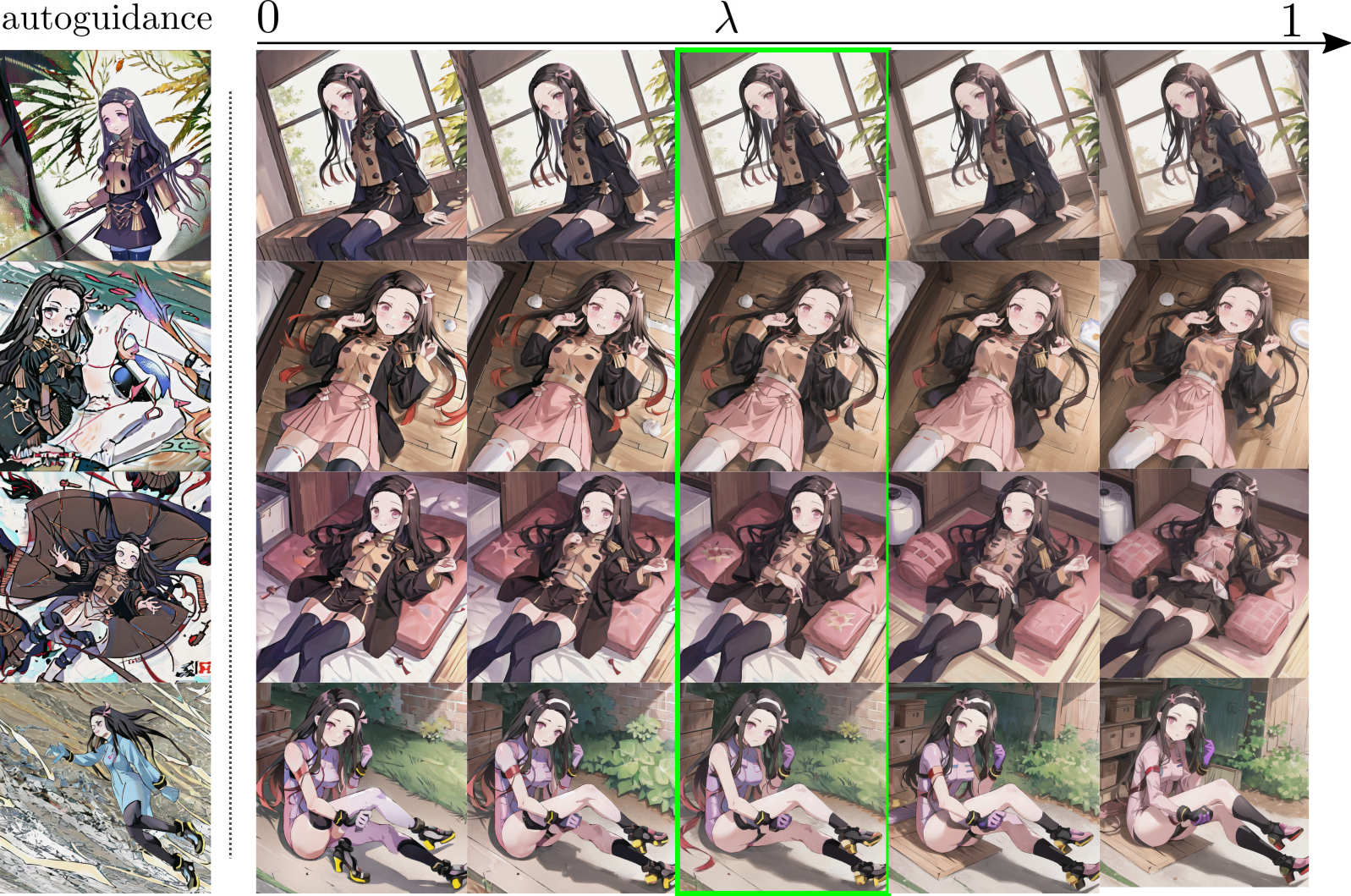}
    \caption[Caption for LOF]{Qualitative comparison of LoRAtorio's re-centering guidance across different values of $\lambda$, evaluated against auto-guidance~\citep{karras_guiding_2024}. The figure illustrates the impact of varying $\lambda$ on image coherence, identity preservation, and visual quality.\footnotemark} 
    \label{fig:rebase_lam}
\end{figure*}

\footnotetext{\clipscore for the images in each row is identical, including autoguidance samples highlighting the necessity of visual inspection and qualitative evaluation.\\
Row 1:32.552, Row 2: 31.810, Row 3: 31.763, Row 4: 32.653
}

\subsection{Ablation Study}\label{app:ablation}
\begin{figure*}[htbp]
    \centering
    \includegraphics[width=\linewidth]{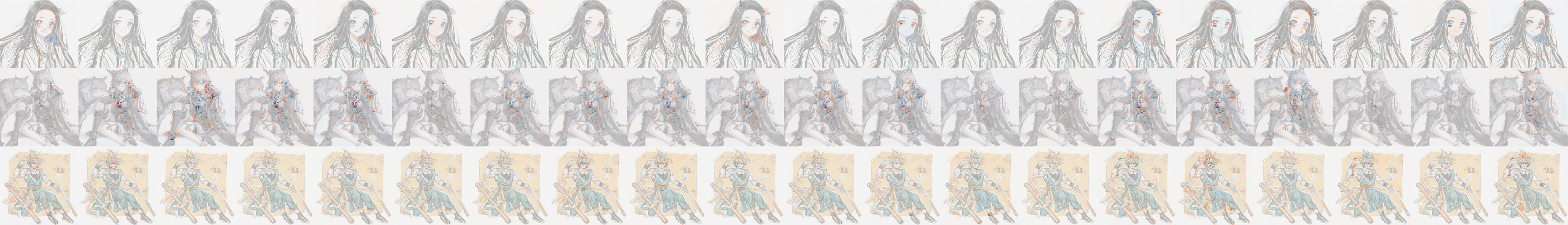}
    \caption{$\Omega^t$ map for character overlayed over the final generated images across timesteps, in Character + Style generation. As style has more global effect, we see clearly the effect of the character LoRA on the predicted noise through the $\Omega^t$ heatmap (high effect is red, no effect is transparent).}
    \label{fig:omega}
\end{figure*}

To show the effect of LoRAtorio's components, we perform an ablation study as shown in~\cref{tab:ablation}, using \clipscore as an evaluation metric. Note that \clipscore is unable to capture the composition of aesthetic quality, so the final set of hyperparameters is selected as a combination of \clipscore and empirically through visual inspection of output.
Specifically, we compute the \clipscore of generated images using the distance of the entire image instead of individual patches, with a constant $\tau=1$ and without our re-centering method. The localised activation of LoRAs through the tokenisation of the latent space has the greatest impact in terms of \clipscore, which is somewhat expected as more elements can be integrated and thus aligned in clip space. We also compare the effect of patch size on the performance of LoRAtorio and see that a more fine-grained composition results in higher \clipscore, although the performance is relatively robust. As such, we select the patch size according to the finest-grained concept in the prompt, since localised concepts (\eg, characters) benefit from finer patches, whereas global LoRAs (\eg, style) are largely insensitive to patch size (\cref{fig:omega}).

\begin{table}[htbp]
\caption{Ablation study of our method on \textit{ComposLoRA} Anime subset, using \clipscore.}\label{tab:ablation}
\centering
\begin{subtable}[t]{0.53\textwidth}
\centering
\caption{LoRAtorio Components}
    \resizebox{\textwidth}{!}{%
        \begin{tabular}{lllll}
        \toprule
                           & $N=2$             & $N=3$         &  Avg.            & \\
        \midrule
        LoRAtorio          & {\ul 36.156}    & \textbf{36.930} & \textbf{36.543} & \\
        w/o $\phi(\cdot)$  & 34.306          & 33.967          & 34.137          & \\
        w/o $\tau$         & 35.948          & {\ul 36.690}    & 36.319          & \\
        w/o Re-centering   & \textbf{36.477} & 36.586          & {\ul 36.532}    & \\
        \bottomrule
        \end{tabular}   
    }
\end{subtable}
\hfill
\begin{subtable}[t]{0.4\textwidth}
\centering
\caption{Patch Size}
    \resizebox{\textwidth}{!}{%
        \begin{tabular}{llll}
        \toprule
                           & $N=2$             & $N=3$         &  Avg.    \\
        \midrule
        $2 \times 2$   & \textbf{36.156} & \textbf{36.930} & \textbf{36.543} \\
        $4 \times 4$   &       36.025    &  36.475             & 36.250 \\
        $8 \times 8$   &       35.852    &  36.423             & 36.138 \\
        $16 \times 16$ &       35.744    &  36.003             & 35.874 \\
        \bottomrule
        \end{tabular}   
    }
\end{subtable}

\end{table}

In addition, we conduct a qualitative comparison of LoRAtorio's re-centering guidance with auto-guidance~\citep{karras_guiding_2024}, and evaluate performance across different values of the weighting parameter $\lambda$, shown on~\cref{fig:rebase_lam}, as \clipscore alone does not capture identity preservation, compositionally or other qualitative elements. More specifically, we show in~\cref{fig:rebase_lam} that the \clipscore is consistent for all selected images, thus reiterating that it should be used as a metric of generic object inclusion, not instance fidelity or qualitative score. Since the base model is essentially ``a bad version'' of the LoRA-augmented model, auto-guidance serves as a natural baseline for assessing our re-centering approach. Notably, we find that combining LoRAtorio with auto-guidance fails to produce coherent images. We hypothesise this is due to a difference in data distribution, a prerequisite for auto-guidance. Similarly, when $\lambda = 0$ -- where the unconditioned score corresponds to that of the base model -- we observe strong identity preservation, but the resulting images exhibit excessive saturation and appear unnatural. Conversely, setting $\lambda = 1$ results in some loss of identity and a blending of concepts. To balance these effects, we empirically select $\lambda = 0.5$ for all experiments, for simplicity. We further corroborate this choice with a pairwise LLM-as-a-judge evaluation over $\lambda \in [0, 1]$ on character, object and clothing concepts, where $\lambda = 0.5$ is preferred in 88\% of comparisons and the adjacent $\lambda = 0.25$ in the remainder.

\begin{figure*}[htbp]
    \centering
    \includegraphics[width=\textwidth]{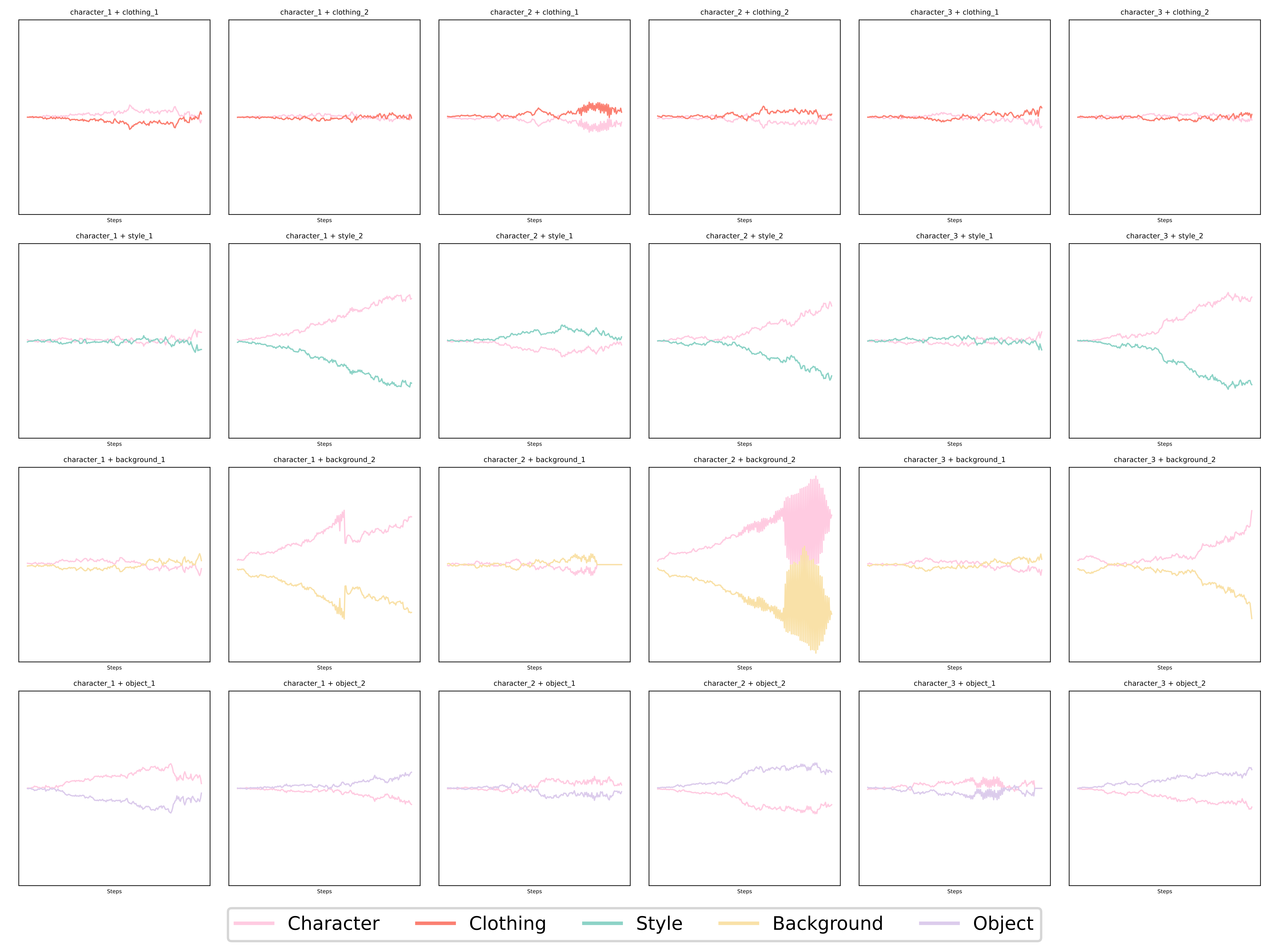}
    \caption{Average values of $\Omega^t$ over denoising process, for the entire image in the \textit{ComposLoRA} testbed for $N=2$.}
    \label{fig:cosine}
\end{figure*}

\section{Temporal Analysis of Similarity-based Weighting}
Our similarity-based weighting mechanism is designed as a proxy of the relative confidence of each LoRA adapter with respect to the base model. Empirically, we observe that the cosine similarity between LoRA-augmented outputs and the base model varies non-uniformly across denoising steps, depending on the LoRA employed.

This temporal asymmetry aligns with prior findings in diffusion literature~\citep{si2024freeu, zhongmulti2024, zou_cached_2025}, which show that different semantic attributes emerge at different stages of the denoising process. However, we observe that the variation within LoRAs of the same type (\eg clothing or style) is too vast for universal and concrete conclusions on the order of activations. We believe this to be due to different LoRA training and configuration, further explored in~\cref{app:limitations}

To validate this behaviour, we analyse the evolution of the similarity matrix $\Omega^t$
  over time, aggregated across the latent representation. As shown in \cref{fig:cosine}, the pattern is not very consistent for any of the element groups. As such, methods relying on guidance based on the type of LoRA used ignore this intrinsic proxy for confidence completely. However, because different elements vary in spatial extent, a na\"{\i}ve global aggregation would disproportionately favour larger elements -- particularly in early steps that affect the trajectory of the denoising process~\citep{zhongmulti2024}. This motivates our use of the spatial tokenisation function $\phi$, which enables fine-grained, patch-level weighting and ensures that larger background or clothing regions do not overshadow smaller but semantically important regions (\eg, smaller objects). A visualisation of the patchwise similarity over the diffusion process can be seen in~\cref{fig:omega}.

\textbf{Spatial Coherence:} To quantify whether independent per-patch weighting introduces spatial fragmentation, we compute the normalised total variation of the $\Omega^t$ maps. We observe that it remains approximately 0.4 throughout the structure-forming window, compared to approximately 1.5 for a fragmented reference, obtained by randomly permuting the same weight values across spatial locations. The weight of each adapter is concentrated in a single coherent region, and the total variation only rises in the late steps as fine texture resolves. Furthermore, as the weighting is applied to the predicted noise in the low-resolution latent space rather than in pixel space, any patch boundaries are smoothed by the remaining denoising steps and the VAE decoder.

\textbf{Region-level Alignment:} We further evaluate whether concepts are routed to their own regions using masked image-text alignment. Specifically, we segment the region of each concept with a grounding model, crop it, and compute its \clipscore against both its own concept phrase and the phrase of a competing concept. We observe that a spatially distinct object concept aligns with its own phrase far more strongly than with the character phrase, with a within-region minus cross-region \clipscore gap of $+9.6$. Note that we report this gap rather than a hard mIoU, as \textit{ComposLoRA} compositions are entangled by construction (\eg, a character wears the clothing and holds the object). As such, a hard mIoU would measure the unavoidable overlap of the benchmark rather than the quality of the gating.

\section{Extended Limitations and Failure Mode Analysis}\label{app:limitations}

\begin{figure*}[htbp]
    \centering
    \includegraphics[width=\textwidth]{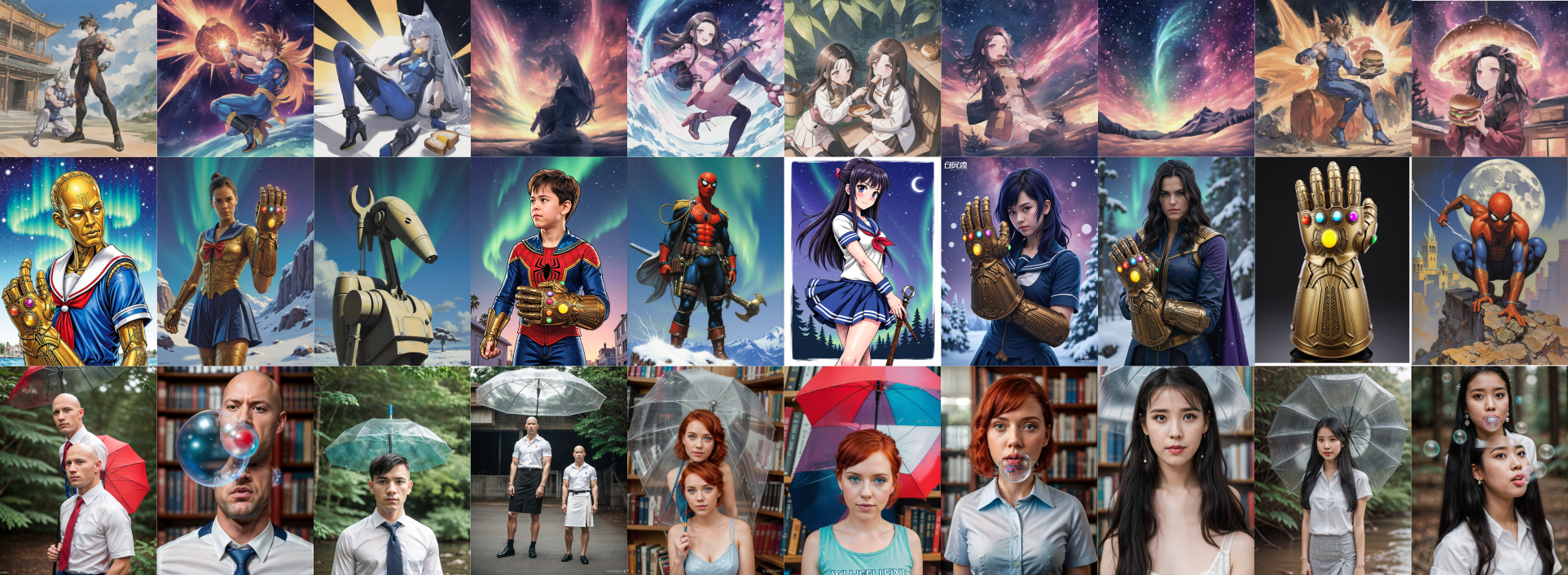}
    \caption{Examples of failure cases of LoRAtorio on \textit{ComposLoRA} anime (top), Flux (mid) and \textit{ComposLoRA} reality (bottom) test beds.}
    \label{fig:failcases}
\end{figure*}

One key limitation of our method is the computational cost. While the intrinsic nature of LoRAtorio allows for a better understanding of the generation process and shows competitive results, the computational cost increases linearly with every additional LoRA. This limitation is identified by previous works manipulating the latent space instead of the weights~\citep{zhongmulti2024}. This is especially true in the open-vocabulary setting where all available LoRA adapters are loaded. Potential future directions to address these limitations include exploring the subspace at an earlier stage (\ie based on early-layer similarity, which we have not explored in the scope of this work), so that pruning or TopK can be implemented before obtaining the latent denoised output. Furthermore, model parallelisation may increase the speed of inference by estimating denoised outputs on different GPUs. 

A primary challenge for any training-free composition method is managing heterogeneous adapter quality. Because LoRAs are trained independently by the community on varying datasets with undocumented hyperparameter configurations, their output magnitudes and latent topologies can differ drastically.

\textbf{Spurious Divergence from Low-Quality LoRAs:} Our core assumption is that latent divergence from the base model acts as a proxy for LoRA relevance and confidence. However, an overfitted or poorly trained LoRA can produce inherently ``noisy'' latent predictions. The SoftMin gating mechanism may misinterpret this chaotic deviation as high-confidence activation, assigning a large weight to a misaligned adapter. This failure mode manifests visually as severe concept confusion or structural collapse.

\textbf{Quantitative Indicators of Failure:} As demonstrated in the quantitative analysis in Section \ref{sec:kl_divergence}, these failure modes can be anticipated by mapping KL-Divergence against CLIP-fidelity. Failures typically occur when a single adapter exhibits an extremely high KL-divergence from the base model but yields a simultaneously low \clipscore. This combination indicates that the LoRA has significantly warped the latent space away from the base distribution without successfully capturing the target semantic concept, leading to inevitable clashes when composed with other adapters.

\textbf{Empirical Validation with Corrupted Adapters:} To validate this indicator, we construct two deliberately corrupted adapters and compare them to five clean \textit{ComposLoRA} adapters. Specifically, we use one over-fitted adapter, trained to memorise a single image, and one domain-mismatched adapter, trained on one concept but registered under the trigger of a different concept. For each adapter, we measure its divergence from the base model and its CLIP-fidelity to the concept it claims. We compute the divergence as the mean relative difference between the noise predictions of the adapter and the base model under the trigger of the adapter, which is an empirical proxy for the KL-divergence above. As shown in~\cref{tab:corrupted}, the over-fitted adapter is identified by its divergence, which is roughly two to three times that of any clean adapter. In contrast, the domain-mismatched adapter falls in the high-divergence, low-fidelity region predicted above. Left unmitigated, SoftMin gating assigns the corrupted adapter the top patch-weight share in 17\% of the probed prompts, and an above-uniform share in 83\%. However, as both corrupted adapters are separable in a single cached forward pass, they can be screened from the candidate pool before composition.

\begin{table}[htbp]
\centering
\caption{Divergence from the base model and CLIP-fidelity for clean and deliberately corrupted adapters.}\label{tab:corrupted}
\begin{tabular}{lcc}
\toprule
Adapter group & Divergence from base & CLIP-fidelity \\
\midrule
Clean (5 adapters) & 0.008--0.022 & 25--36 \\
Over-fitted & 0.050 & 35 \\
Domain-mismatched & 0.082 & 16.6 \\
\bottomrule
\end{tabular}
\end{table}

Our method assumes that LoRA adapters are trained on reasonably well-aligned and semantically coherent datasets. In practice, however, LoRA quality can vary significantly, particularly when sourced from community repositories such as CivitAI, where training data, objectives, and preprocessing pipelines are often undocumented or inconsistent. This variability can undermine the reliability of any train-free approach. Moreover, the LoRAs used during inference are heterogeneous in terms of optimal hyperparameters (\eg, guidance scale, LoRA scale), and treating them uniformly may inadvertently bias the composition toward certain adapters, especially those with more aggressive or dominant activations. While we expect better performance when LoRAs are trained under similar conditions, such alignment is rarely guaranteed in user-driven settings. One potential mitigation strategy in real-life applications is to incorporate a lightweight pre-filtering step to assess LoRA quality before inclusion. Alternatively, metadata-based heuristics (\eg, dataset size, training steps, or CFG guidance scale) could be used to cluster or filter LoRAs. Although these approaches are not explored in this work beyond the divergence--fidelity screening above, they represent promising directions for improving robustness in real-world deployments. Finally, as all LoRAs in the \textit{ComposLoRA} and Flux testbeds are sourced from CivitAI without access to training details, we emphasise that all results should be interpreted in light of this uncertainty.

Finally, we note that the quality of images is affected by the base model.~\cref{fig:failcases} shows examples of fail cases of LoRAtorio for all three base models. We observe that the Stable Diffusion backbone (top and bottom rows) exhibits more instances of additional limbs or duplicate characters compared to the Flux backbone (middle row), where most failure cases are attributed to concept confusion. As such, expanding the test bed to more backbones is essential in dissecting base model vs method limitations.

\section{Flux Testbed and Implementation Details}\label{app:fluxlora}

For experiments on Flux, we select the LoRAs described in~\cref{tab:fluxlora}, following a selection process similar to \textit{ComposLoRA}. All LoRAs used in the Flux experiments are publicly available through CivitAI. We select LoRAs for three characters, two clothing items, two styles, two objects and one background.

\begin{table*}[htbp]
\caption{Details of LoRA adapters used in Flux experiments.}\label{tab:fluxlora}
\resizebox{\textwidth}{!}{%
\begin{tabular}{llll}
    \toprule
    LoRA                                 & Category   & Trigger                                                           & Source                                                                                                              \\
    \midrule
    Yennefer of Vengerberg               & Character  & Yennefer                                                          & \href{https://civitai.com/models/120703/yennefer-witcher-3-game-fluxsdxl}{Link} \\
    The amazing Spiderman                & Character  & Spider-Man, Peter Parker                                          & \href{https://civitai.com/models/196131/the-amazing-spider-man-xl-sd15-f1d-illu-pony}{Link}                             \\
    B1 Battle Droid                      & Character  & 7-B1 droid                                                        & \href{https://civitai.com/models/361684/star-wars-b1-battle-droid-fluxsdxlsd15pony}{Link}                              \\
    \midrule
    Star Wars imperial officer uniform   & Clothing   & Wearing an imperial officer IMPOFF uniform                        & \href{https://civitai.com/models/28801/star-wars-imperial-officer-uniform-flux-and-sd-15}{Link}                        \\
    Japanese school uniform - sailorfuku & Clothing   & wearing a japanese school uniform sailorfuku serafuku sailor suit & \href{https://civitai.com/models/922785/japanese-school-uniform-sailorfuku}{Link}                                     \\
    \midrule
    Frank Frazetta Style Oil Painting    & Style      & in the style of Frank Frazetta fantasy oil painting               & \href{https://civitai.com/models/657789/flux-frank-frazetta-style-oil-painting}{Link}                                           \\
    Engraving Style                      & Style      & in engraving style                                                & \href{https://civitai.com/models/758278/flux-engraving-style-pen-and-ink-drawing-style}{Link}                                      \\
    \midrule
    Infinity Goblet                      & Object     & with a glove like Infinity Gauntlet                               & \href{https://civitai.com/models/904739/the-infinity-gauntlet-flux1d}{Link}                                                   \\
    Crescent Wrench                      & Object     & with a Crescent Wrench                                            & \href{https://civitai.com/models/313711/crescent-wrench}{Link}                                                       \\
    \midrule
    Northern Lights                      & Background & with Northern lights style background                             & \href{https://civitai.com/models/1269331/better-northern-lights-photography-for-flux-or-realistic-northern-lights-auroras}{Link} \\
    \bottomrule
\end{tabular}
}
\end{table*}

\subsection{Architecture Adaptations and Baseline Selection}
To our knowledge, we are the first to extend multi-LoRA inference-time composition to rectified-flow architectures, specifically targeting the transformer-based Flux.1-dev model. This required specific adaptations to the LoRAtorio framework:

\textbf{Omission of Tokenization and CFG Re-centering:} Unlike Stable Diffusion v1.5, Flux relies on a transformer-based architecture where the latent space is natively tokenised; therefore, we apply our weighting directly to the denoised network outputs without requiring an explicit patch-tokenization step. Furthermore, Flux does not employ traditional Classifier-Free Guidance (CFG)—where inputs are passed twice through the network—but instead embeds control signal parameters directly within the cross-attentions. Consequently, the CFG adjustment described in Equations 1 and 2 is structurally incompatible with Flux, and the re-centering step is naturally omitted. Equations 3 through 6 remain fully applicable.

\textbf{Baseline Selection:} Adapting existing multi-LoRA methods to rectified-flow architectures requires extensive re-engineering. We selected Na\"{\i}ve composition and LoRA Merge as our primary baselines for the Flux testbed because they are the only training-free, multi-LoRA methods natively supported out-of-the-box for this architecture.

\begin{figure*}[htbp]
    \begin{lstlisting}[language=Python]
def evaluate():
    image_n = 196 # number of images to evaluate
    gpt4v = GPT4V()
    # Load images
    image_path = "images"
    images = []
    for i in range(0, image_n + 0):
        cur_image = encode_image(join(image_path, f"{i}.png"))
        images.append(cur_image)
    
    # Load prompts usd to generate the images
    prompts = []
    with open("image_info.json") as f:
        image_info = json.loads(f.read())
    for i in range(len(image_info)):
        cur_prompt = "\n".join(image_info[i]["prompt"])
        prompts.append(cur_prompt)

    # Comparative evaluation
    gpt4v = GPT4V()
    gpt4v_scores = [{} for _ in range(image_n)]
    # i:     method 1
    # i + 1: method 2
    for i in tqdm(range(0, image_n, 2)):
        method1_image     = images[i]
        method2_image    = images[i + 1]

        cur_prompt = get_eval_prompt(prompts[i])
        
        compare_images(method1_image, method2_image, "method_1", "method_2", gpt4v, cur_prompt)
        compare_images(method2_image, method1_image, "method_2", "method_1", gpt4v, cur_prompt)
\end{lstlisting}
    \caption{Pseudo-code of~\gpt Evaluation. For each pair of images, the comparison is run twice to account for bias in presentation order.}
    \label{fig:code}
\end{figure*}

\section{\gpt Evaluation Interface}
For the~\gpt evaluation, we use the prompt and method of the original \textit{ComposLoRA}~\cite{zhongmulti2024}. Specifically, we do pairwise comparisons of our method against previous works. The comparison is run twice for each pair, switching the order of images to account for any bias induced from ordering. The scores are then averaged. Pseudo-code of the evaluation can be seen in~\cref{fig:code}. The prompt used in the evaluation can be seen in~\cref{sec:prompt}.

% \subsection{GPT4v Prompt}\label{sec:prompt}

\begin{figure*}[htbp]
    \begin{lstlisting}
I need assistance in comparatively evaluating two text-to-image models based on their ability to compose different elements into a single image. The elements and their key features are as follows:
    <IMAGE_1> <IMAGE_2> <PROMPT>

Please help me rate both given images on the following evaluation dimensions and criteria:

Composition quality:
    - Score on a scale of 0 to 10, in 0.5 increments, where 10 is the best and 0 is the worst.
    - Deduct 3 points if any element is missing or incorrectly depicted.
    - Deduct 1 point for each missing or incorrect feature within an element.
    - Deduct 1 point for minor inconsistencies or lack of harmony between elements.
    - Additional deductions can be made for compositions that lack coherence, creativity, or realism.


Image quality:
    - Score on a scale of 0 to 10, in 0.5 increments, where 10 is the best and 0 is the worst.
    - Deduct 3 points for each deformity in the image (e.g., extra limbs or fingers, distorted face, incorrect proportions).
    - Deduct 2 points for noticeable issues with texture, lighting, or color.
    - Deduct 1 point for each minor flaw or imperfection.
    - Additional deductions can be made for any issues affecting the overall aesthetic or clarity of the image.


Please format the evaluation as follows:

For Image 1:

[Explanation of evaluation]

For Image 2:

[Explanation of evaluation]

Scores:

Image 1: Composition Quality: [score]/10, Image Quality: [score]/10

Image 2: Composition Quality: [score]/10, Image Quality: [score]/10

Based on the above guidelines, help me to conduct a step-by-step comparative evaluation of the given images. The scoring should follow two principles:
    1. Please evaluate critically.
    2. Try not to let the two models end in a tie on both dimensions.

\end{lstlisting}
\caption{\gpt prompt}
\label{sec:prompt}
\end{figure*}

\begin{table}[htbp]
\centering
\caption{Comparison of Multiply-Accumulate Operations (MACs) and qualitative latency estimates under different $N$ values.}
\label{tab:macs_latency_comparison}
\begin{subtable}[t]{0.48\textwidth}
\centering
\caption{MACs (in GigaOps) for different methods.}\label{tab:macs_comparison}
\resizebox{\textwidth}{!}{%

\begin{tabular}{lcccc}
\toprule
              & $N=2$       & $N=3$      & $N=4$      & $N=5$      \\
\midrule
LoRAtorio     & $1090.863$  & $1102.721$ & $1125.570$ & $1132.123$ \\
CMLoRA$^a$    & $912.350$   & $1223.486$ & $1358.518$ & $1570.335$ \\
Switch-A$^a$  & $734.053$   & $730.914$  & $739.322$  & $731.811$  \\
LoraHub$^a$   & $789.770$   & $834.613$  & $924.299$  & $946.721$  \\
Composite$^a$ & $1401.066$  & $2169.199$ & $2892.266$ & $3615.333$ \\
Switch$^a$    & $734.053$   & $730.914$  & $739.322$  & $731.811$  \\
Merge$^a$     & $789.770$   & $834.613$  & $924.299$  & $946.721$  \\
\bottomrule
\end{tabular}
}
\end{subtable}
\hfill
\begin{subtable}[t]{0.48\textwidth}
\centering
\caption{Qualitative latency estimates (seconds).}\label{tab:latency_comparison}
\resizebox{\textwidth}{!}{%

\begin{tabular}{lcccc}
\toprule
              & $N=2$ & $N=3$ & $N=4$ & $N=5$ \\
\midrule
LoRAtorio     & 61    & 85    & 91    & 122   \\
Merge$^b$     & 20    & 21    & 22    & 24    \\
Switch$^b$    & 16    & 18    & 19    & 20    \\
Composite$^b$ & 60    & 70    & 76    & 90    \\
MultLFG$^b$   & 90    & 140   & 180   & 230   \\
\bottomrule
\end{tabular}
}
\end{subtable}

\vspace{0.5em}
\footnotesize{$^a$ As reported by~\cite{zou_cached_2025}}

\footnotesize{$^b$ As reported by~\cite{roy2025multlfg}}

\end{table}

\section{Computational Cost Analysis and Dynamic Top-$k$ Selection}\label{app:cost}

In the dynamic module selection setting, LoRAtorio is tasked with selecting the most relevant adapters from a large pool without prior knowledge of the prompt's requirements. 

\textbf{Top-$k$ Configuration:} We clarify that the top-$k$ selection operates locally on the $\Omega^t$ matrix, which has dimensionality $N \times P$ (where $N$ is the number of loaded LoRAs and $P$ is the number of spatial patches). For all dynamic experiments, we explicitly set $k=C$, where $C$ is the number of distinct semantic concepts requested in the text prompt, allowing for a fair comparison against the static setting. 

\textbf{Computational Cost:} Achieving true zero-shot dynamic selection requires evaluating the denoising network for all $N$ loaded adapters at each timestep to compute the similarity matrix $\Omega^t$ and rank the outputs. While this increases the per-step memory footprint and computational overhead linearly with respect to $N$, \cref{tab:macs_latency_comparison} demonstrates that the total Multiply-Accumulate Operations (MACs) and inference latency remain highly competitive.

While LoRAtorio demonstrates competitive performance and interpretability, its computational cost increases linearly with the number of active LoRA modules. This is a direct consequence of its design, which composes multiple LoRAs simultaneously in the latent space. In contrast, methods like Switch or Merge maintain a relatively constant cost by activating only a subset or a merged representation of LoRAs. This limitation aligns with prior observations in latent-space manipulation approaches~\citep{zhongmulti2024}. The cost becomes particularly significant in dynamic settings, where all available LoRA adapters may be loaded concurrently. However, we also note that the MACs of LoRAtorio do not differ by orders of magnitude compared to previous works; in fact, we see that they are comparable, thus our method does not introduce a significant cost-performance trade-off compared to previous works. In~\cref{tab:latency_comparison}, we also see a comparison of LoRAtorio against reported inference latency in seconds. Our method has comparable latency to LoRA-composite and is significantly faster than MultLFG. Furthermore, restricting the gating over all $N$ candidates to an early window of $K$ steps reduces the dynamic cost from $\mathcal{O}(N \cdot T)$ towards $\mathcal{O}(N \cdot K + C \cdot T)$, and the per-adapter passes can be parallelised across devices. For web-scale adapter libraries, we expect a retrieval-stage pre-selection (\eg, based on metadata) to be applied before LoRAtorio, which then composes over the loaded candidate set.

\textbf{Exploratory Threshold-based Selection:} We further explore a zero-shot variant that does not assume a known $k$. Specifically, we load all candidate adapters and accumulate the aggregate gating signal of each adapter over a short window of denoising steps, \ie, the extent to which its predicted noise diverges from the base model across patches. We compute the same signal under the empty prompt as a per-adapter baseline, with the base model acting as a null adapter. We then subtract this baseline and retain every adapter whose calibrated signal exceeds a single global margin $\delta$. The margin is fixed once on a small calibration split and applied unchanged to all prompts, so that the number of concepts $C$ is never provided. We evaluate selection directly as retrieval against the ground-truth concept set of each prompt, and compare against a known-$k$ oracle that selects the top-$C$ adapters by the same signal, as well as an always-all-$N$ baseline that retains every loaded adapter. As shown in~\cref{tab:kfree_prelim}, the threshold rule matches the known-$k$ oracle on a preliminary \textit{ComposLoRA} evaluation, without access to $C$. When the entire adapter pool is loaded, the threshold rule exceeds both baselines in F1 for two- and three-concept compositions (\cref{tab:kfree_full}). We note that its high recall ensures that relevant concepts are rarely dropped, while admitted irrelevant adapters are suppressed by the subsequent per-patch SoftMin gating.

\begin{table}[htbp]
\centering
\caption{Exploratory retrieval results for threshold-based (k-free) dynamic selection.}
\label{tab:kfree}
\begin{subtable}[t]{0.48\textwidth}
\centering
\caption{Preliminary \textit{ComposLoRA} evaluation.}\label{tab:kfree_prelim}
\resizebox{\textwidth}{!}{%
\begin{tabular}{lccc}
\toprule
Rule & Precision & Recall & F1 \\
\midrule
Threshold (no $C$) & 0.75 & 0.75 & 0.75 \\
Known-$k$ oracle (top-$C$) & 0.75 & 0.75 & 0.75 \\
Always-all-$N$ & 0.50 & 1.00 & 0.67 \\
\bottomrule
\end{tabular}
}
\end{subtable}
\hfill
\begin{subtable}[t]{0.48\textwidth}
\centering
\caption{Full adapter pool loaded.}\label{tab:kfree_full}
\resizebox{\textwidth}{!}{%
\begin{tabular}{lccc}
\toprule
Rule & Precision & Recall & F1 \\
\midrule
Threshold (no $C$) & 0.40 & 0.83 & \textbf{0.54} \\
Known-$k$ oracle (top-$C$) & 0.25 & 0.25 & 0.25 \\
Always-all-$N$ & 0.23 & 1.00 & 0.37 \\
\bottomrule
\end{tabular}
}
\end{subtable}
\end{table}

\section{Qualitative Comparison}\label{app:qual}

\begin{figure*}[htbp]
        \centering
        \includegraphics[width=\textwidth]{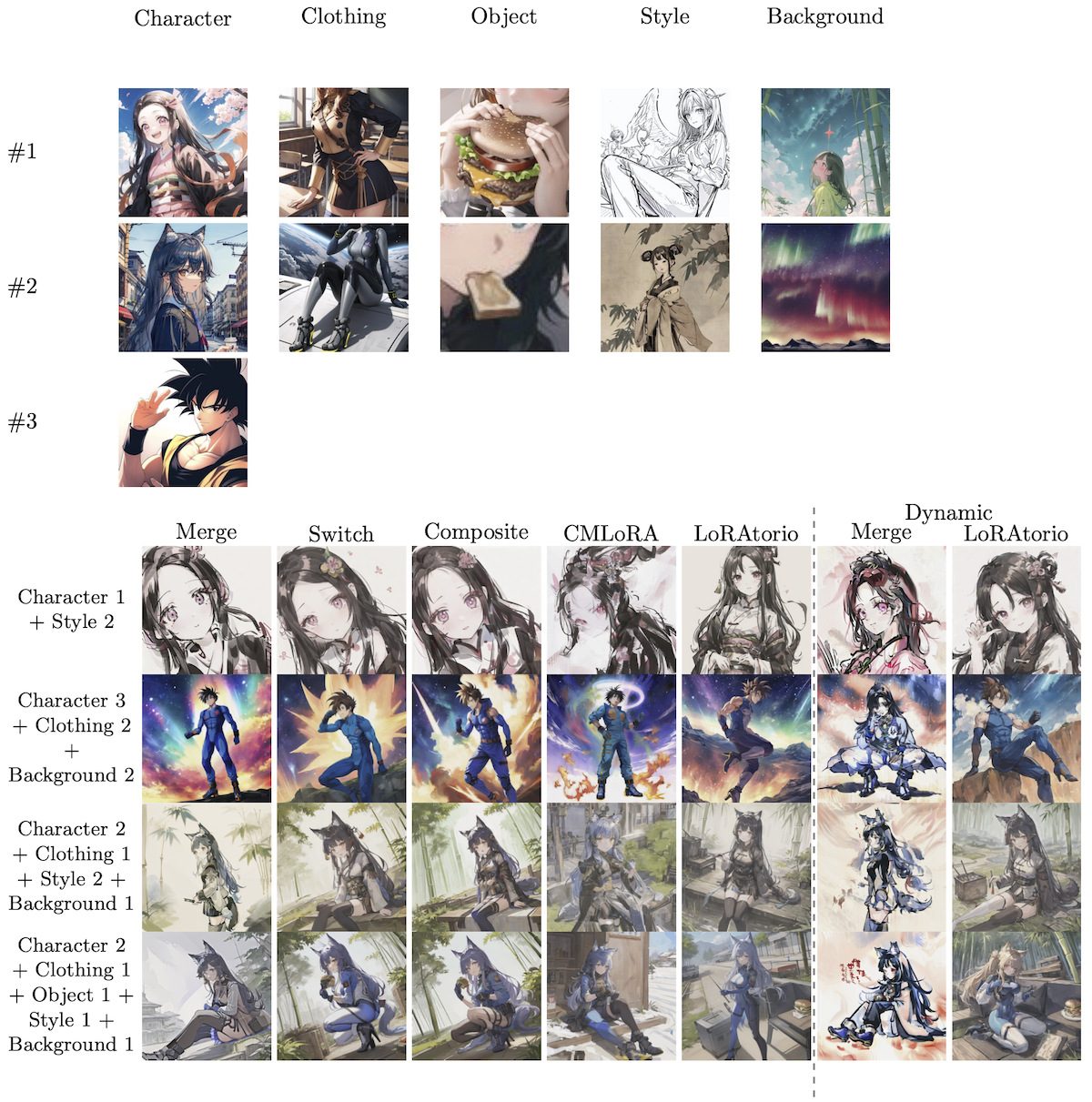}
        \caption{Images generated with $N$ LoRA candidates (L1 Character, L2 Clothing, L3 Style, L4 Background and L5 Object) across our proposed framework and baseline methods using SD1.5 base model on the anime subset of \textit{ComposLoRA}.}
        \label{fig:sd1_anime}
    \end{figure*}

\begin{figure*}[htbp]
        \centering
        \includegraphics[width=\textwidth]{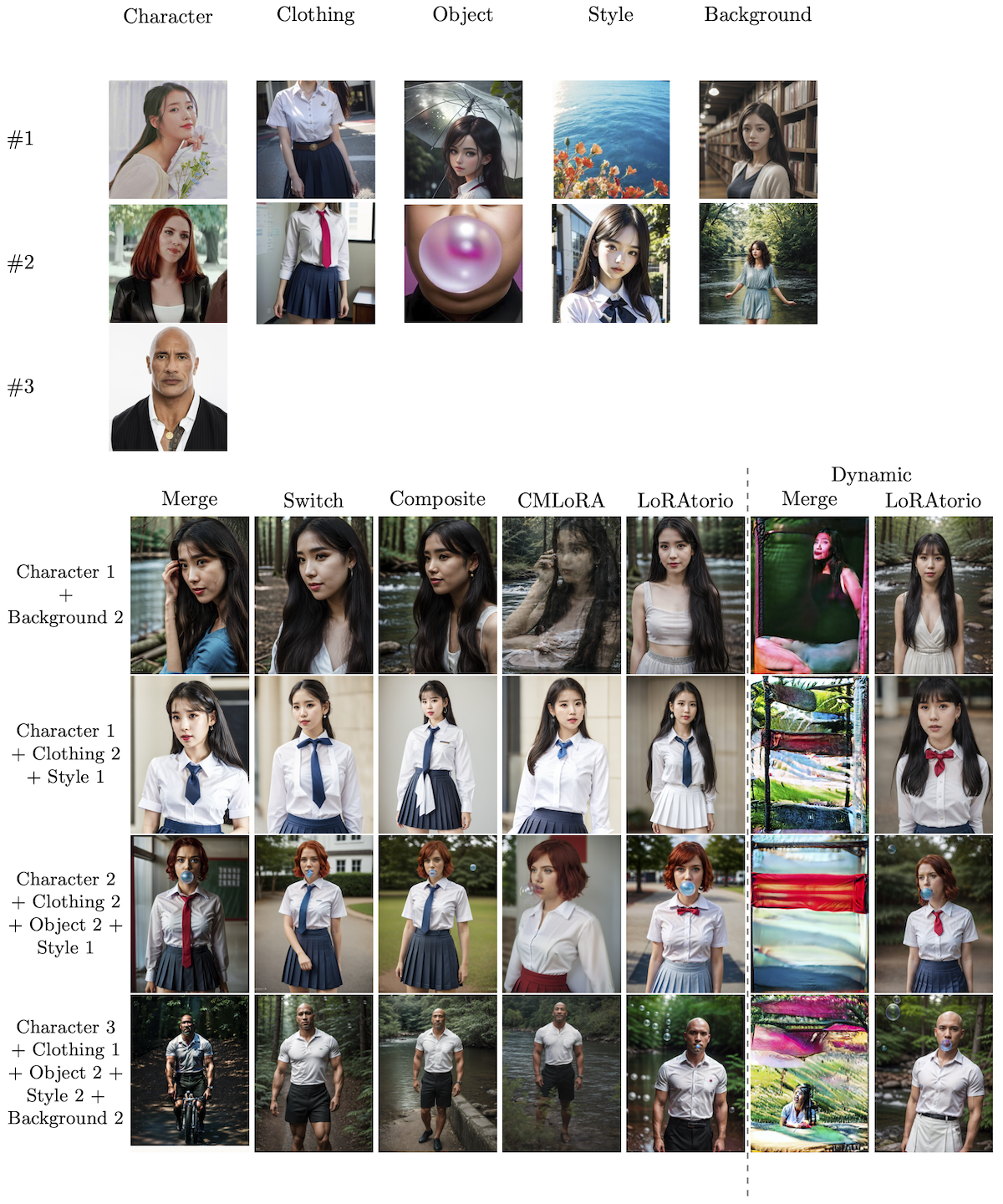}
        \caption{Images generated with $N$ LoRA candidates (L1 Character, L2 Clothing, L3 Style, L4 Background and L5 Object) across our proposed framework and baseline methods using SD1.5 base model on the reality subset of \textit{ComposLoRA}.}
        \label{fig:sd1_reality}
    \end{figure*}

Examples comparing our method against baselines qualitatively for the SD1.5 base model can be seen in~\cref{fig:sd1_anime} and~\cref{fig:sd1_reality}. LoRAtorio performs competitively, exhibiting fewer concept clashes and reduced vanishing of key attributes compared to previous SoTA. In the dynamic selection setting, we observe that Merge collapses, often producing non-legible or incoherent images, especially in the more complex reality subset. In contrast, LoRAtorio reliably selects the most relevant LoRA modules, resulting in sharper, more coherent generations, with clearly recognisable concepts.

\begin{figure*}[htbp]
        \centering
        \includegraphics[width=\textwidth]{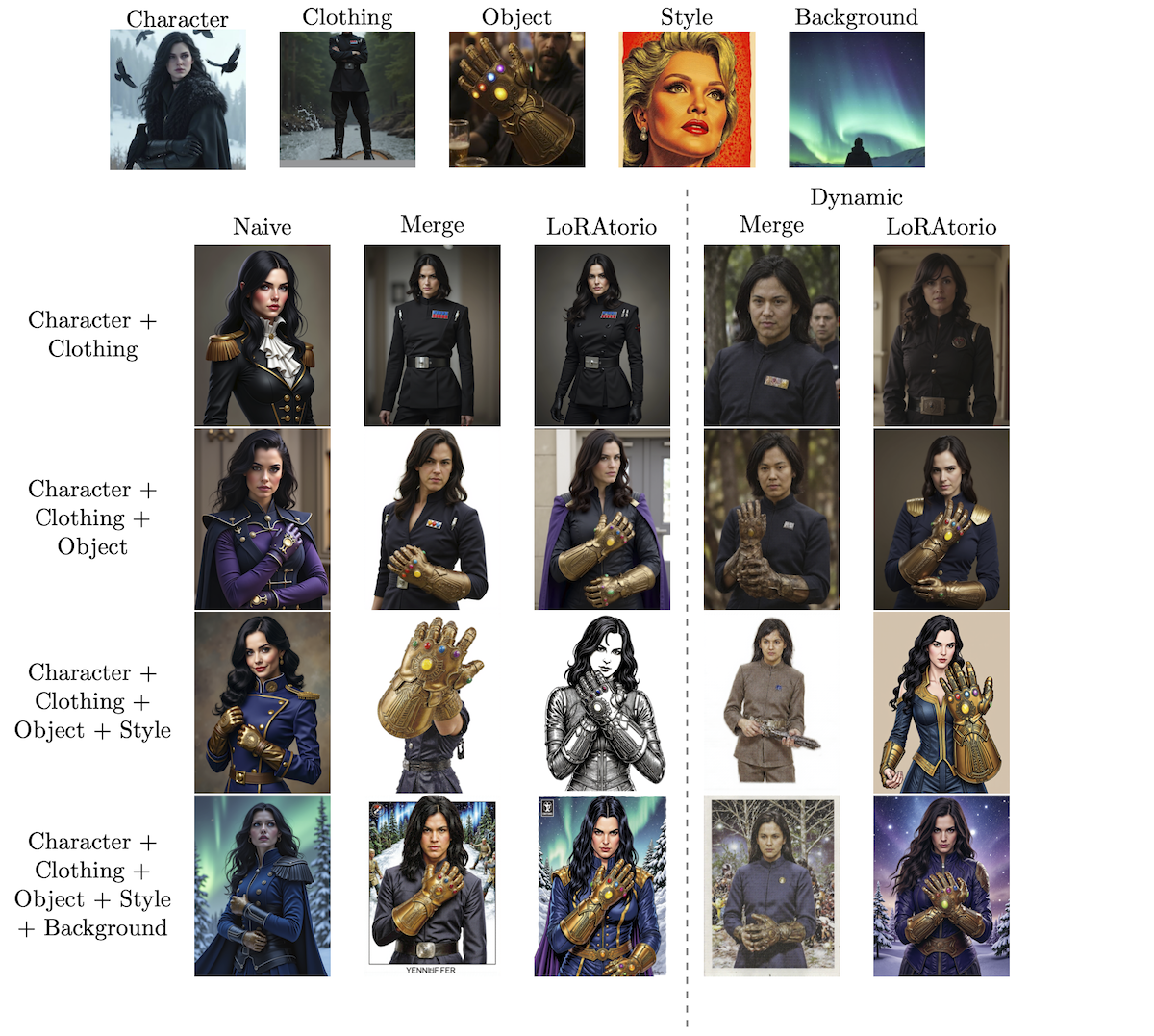}
        \caption{Images generated with $N$ LoRA candidates (L1 Character, L2 Clothing, L3 Style, L4 Background and L5 Object) across our proposed framework and baseline methods using Flux base model.}
    \label{fig:flux_char1}
    \end{figure*}

\begin{figure*}[htbp]
        \centering
        \includegraphics[width=\textwidth]{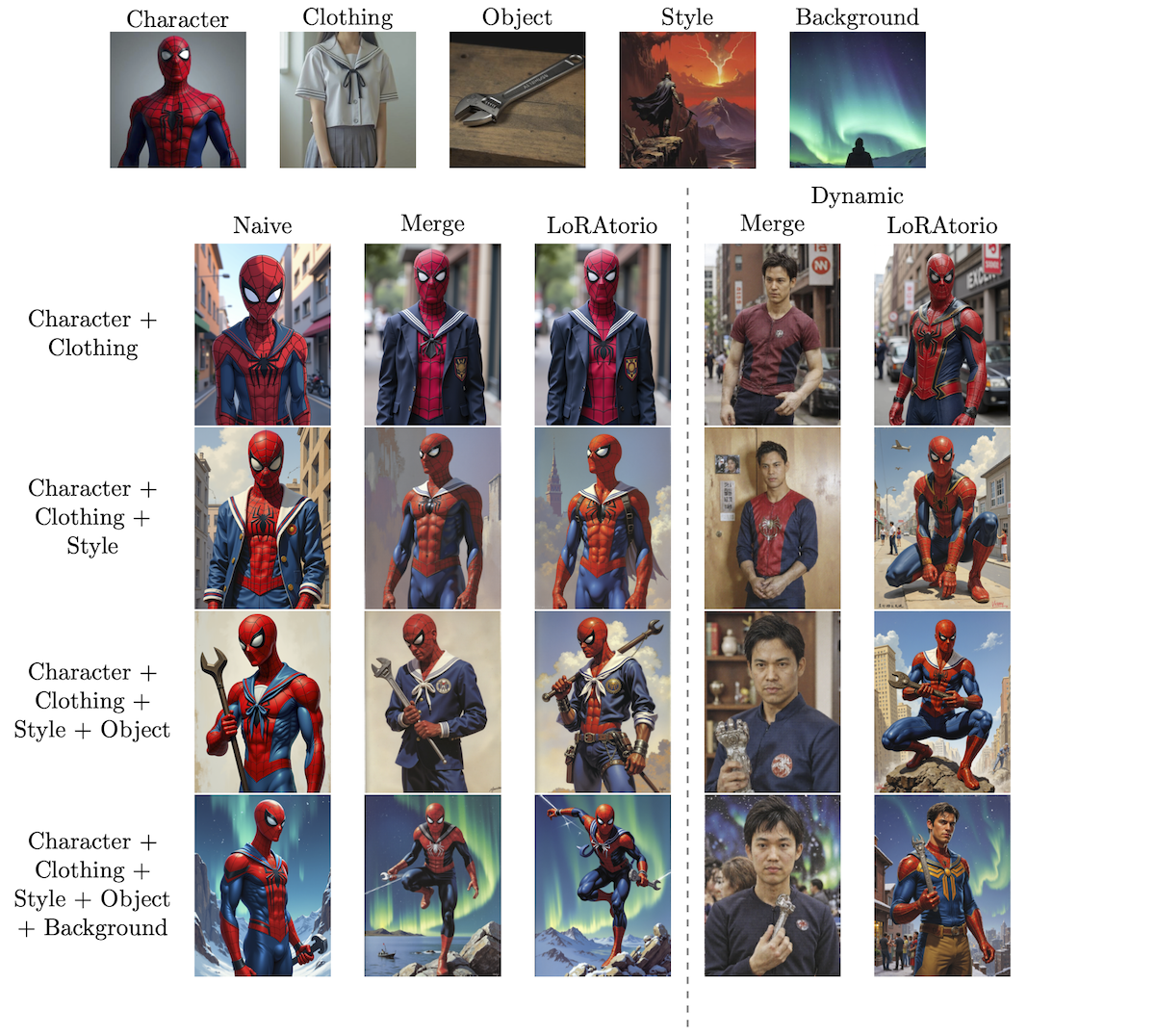}
        \caption{Images generated with $N$ LoRA candidates (L1 Character, L2 Clothing, L3 Style, L4 Background and L5 Object) across our proposed framework and baseline methods using Flux base model.}
    \label{fig:flux_char2}
    \end{figure*}

\begin{figure*}[htbp]
        \centering
        \includegraphics[width=\textwidth]{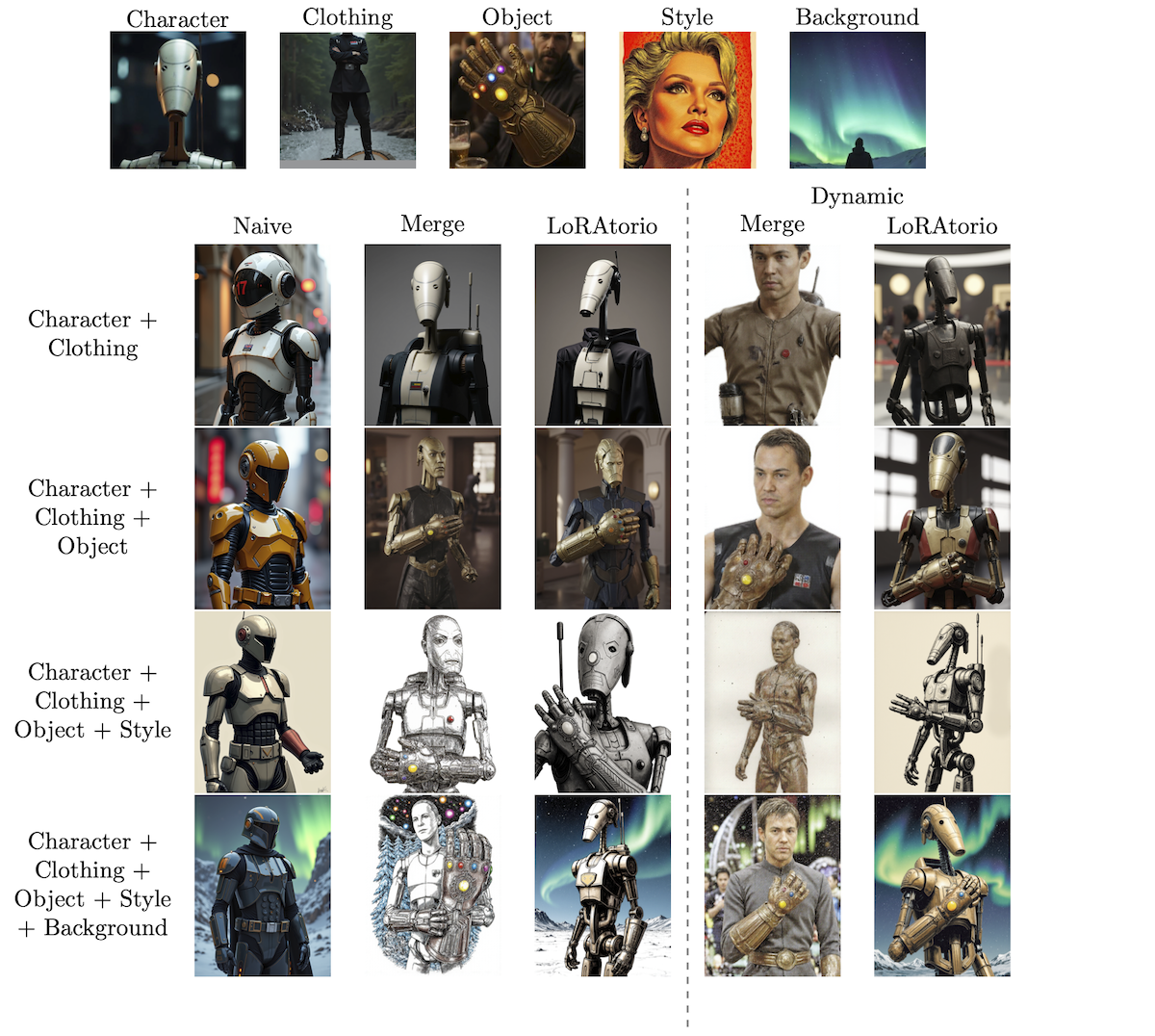}
        \caption{Images generated with $N$ LoRA candidates (L1 Character, L2 Clothing, L3 Style, L4 Background and L5 Object) across our proposed framework and baseline methods using Flux base model.}
    \label{fig:flux_char3}
    \end{figure*}

Examples comparing our method against baselines for the Flux base model can be seen in~\cref{fig:flux_char1},~\cref{fig:flux_char2}, and~\cref{fig:flux_char3}. LoRAtorio shows lower concept confusion compared to merge. This is particularly obvious in the case of the Dynamic module selection setting, where the image quality severely deteriorates with multiple LoRAs. Even though Flux is a stronger model and thus generates more legible images compared to SD1.5 in the dynamic setting, the difference in quality compared to LoRAtorio and Na\"{\i}ve is substantial. 

    % \subimport{figures/}{fig_fluxexamples}
    
\clearpage
\section{Human Evaluation Interface}\label{app:human}
For the human qualitative evaluation, we follow the framework of~\cite{zou_cached_2025} along four qualitative axes. For each combination, we provide a set of reference images and the output of the anonymised methods. Samples of the instructions and survey, as shown to human experts, are presented below. We use three human experts to evaluate our method against previous SoTA. The annotators have an average Cohen's Kappa of $0.67$, indicating agreement.
\includepdf[pages=-, scale=0.8, pagecommand={\thispagestyle{plain}}]{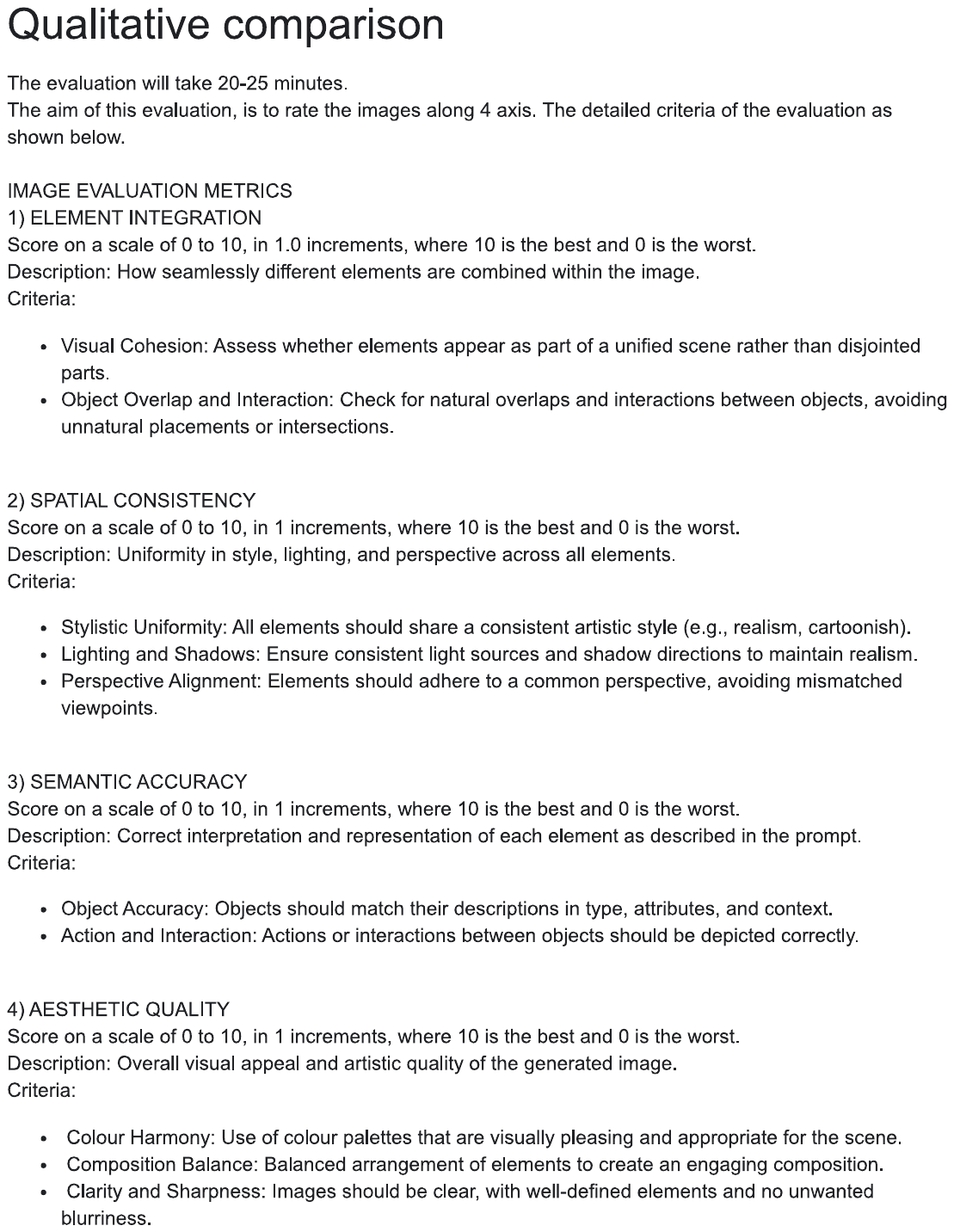}
\includepdf[pages=-, scale=0.8, pagecommand={\thispagestyle{plain}}]{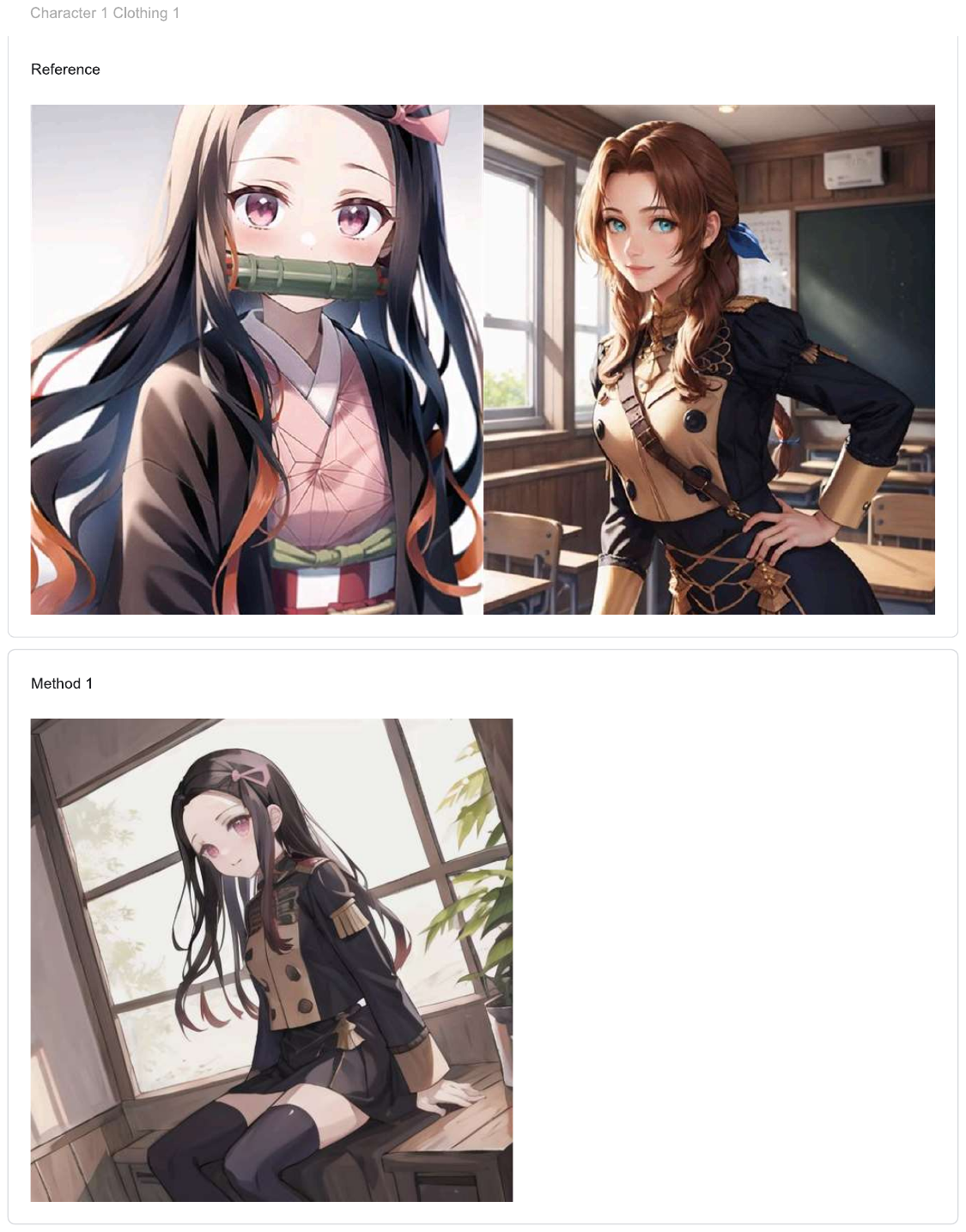}
%%%%%%%%%%%%%%%%%%%%%%%%%%%%%%%%%%%%%%%%%%%%%%%%%%%%%%%%%%%%

\end{document}